\documentclass[10pt,letterpaper,twocolumn,teaser,logo]{planarxiv}
\usepackage{xspace}
\usepackage{mathtools,mathrsfs}
\usepackage{pifont,rotating}
\usepackage{adjustbox}
\usepackage{pgfplots}
\pgfplotsset{compat=1.18}
\usepgfplotslibrary{groupplots,fillbetween}
\usepackage{gradient-text}
\IfFileExists{dsfont.sty}{\usepackage{dsfont}}{}
\IfFileExists{bbm.sty}{\usepackage{bbm}}{}

\crefname{equation}{Eq.}{Eqs.}
\crefformat{section}{\S#2#1#3}
\crefformat{subsection}{\S#2#1#3}
\crefformat{subsubsection}{\S#2#1#3}
\crefrangeformat{section}{\S\S#3#1#4 to~#5#2#6}
\crefmultiformat{section}{\S\S#2#1#3}{ and~#2#1#3}{, #2#1#3}{ and~#2#1#3}

\DeclareMathOperator*{\E}{\mathbb{E}}                   
\newcommand{\mc}{\mathcal}        

\newcommand{\loss}{\mathcal{L}}              
\newcommand{\wh}{\widehat}                  

\newcommand{\eng}{\mc{G}}                      
\newcommand{\crit}{\mc{C}}                     
\newcommand{\pol}{\pi_{\theta}}                
\newcommand{\Qf}{Q_{\phi}}                     

\newcommand{\env}{s}                           
\newcommand{\prog}{g}                          
\newcommand{\vfy}{\nu}                         
\newcommand{\tsk}{\tau}                        
\newcommand{\spa}{\sigma}                      
\newcommand{\att}{\alpha}                      
\newcommand{\Exec}{\operatorname{Exec}}
\newcommand{\sdk}[1]{\texttt{#1}}

\newcommand{\attr}{\psi}                       
\newcommand{\axis}{\delta}                     
\newcommand{\taskax}{\textsc{task}\xspace}
\newcommand{\spatax}{\textsc{spatial}\xspace}
\newcommand{\attax}{\textsc{attribute}\xspace}
\newcommand{\execax}{\textsc{execution}\xspace}

\newcommand{\lv}{\operatorname{LV}}            
\newcommand{\psucc}{\wh{p}}                    

\newcommand{\mem}{\mc{M}}                      
\newcommand{\buf}{\mc{B}}                      
\newcommand{\skpool}{\Omega}                   
\newcommand{\dirpool}{\Phi}                    
\newcommand{\clus}{\mc{K}}                     
\newcommand{\Rgrind}{\mc{R}_{\mathrm{grind}}}  
\newcommand{\Rhone}{\mc{R}_{\mathrm{hone}}}    

\newcommand{\dem}{\mc{D}}                      

\newcommand{\pot}{\Phi}                        
\newcommand{\potsub}[1]{\varphi_{#1}}          
\newcommand{\chunk}[1]{a_{#1:#1+h}}            
\newcommand{\shaped}{\tilde{r}}                
\newcommand{\subg}{\mathrm{g}}                 

\newcommand{\mempot}{\textbf{Text2potential}\xspace}

\usepackage[nospread,noshrink]{flushend}

\definecolor{saprE}{RGB}{201, 182, 228} 
\definecolor{rlE}{RGB}{245,130,55}

\definecolor{fire0}{HTML}{FFF2B2}
\definecolor{fire1}{HTML}{C1121F}
\definecolor{fire2}{HTML}{FFB347}
\definecolor{fire3}{HTML}{FF8A3D}
\definecolor{fire4}{HTML}{FF4D00}
\definecolor{fire5}{HTML}{FF4A3A}
\definecolor{fire6}{HTML}{CE0A18}

\definecolor{CUBlue}{HTML}{4285F4}
\definecolor{Google}{HTML}{EA4335}

\definecolor{pyraAmber}{HTML}{FFB347}
\definecolor{pyraOrange}{HTML}{FF6B3D}
\definecolor{pyraRed}{HTML}{E63946}

\definecolor{color1}{RGB}{220, 237, 220}  
\definecolor{color2}{RGB}{252, 224, 225}  
\definecolor{color3}{RGB}{220, 235, 247}  
\definecolor{color4}{RGB}{230, 230, 230}  

\theoremstyle{plain}

\theoremstyle{definition}

\theoremstyle{remark}

\usepackage{enumitem}
\setlist[itemize]{leftmargin=*}

\definecolor{IllinoisBlue}{HTML}{13294B}
\definecolor{IllinoisOrange}{HTML}{FF5F05}
\definecolor{SoftBeige}{HTML}{FEF5E7}
\definecolor{YaleBlue}{HTML}{2A5487}
\definecolor{CustomRed}{RGB}{233, 93, 34}      
\definecolor{CustomOrange}{RGB}{245, 132, 38}  
\definecolor{CustomPeach}{RGB}{255, 165, 82}   
\definecolor{CustomLightOrange}{RGB}{255, 170, 120} 
\definecolor{CustomLightRedOrange}{RGB}{255, 195, 160} 
\definecolor{CustomYellow}{RGB}{255, 220, 148} 
\definecolor{CustomDarkGray}{RGB}{80, 80, 80}
\definecolor{CustomTeal}{HTML}{008080}   
\definecolor{BrightTeal}{HTML}{00A6A6}
\definecolor{CustomMint}{HTML}{50C99A}
\definecolor{CustomPink}{HTML}{FF828F}
\definecolor{CustomPurple}{HTML}{884AB2}
\definecolor{CustomLightPurple}{HTML}{C497FF}
\definecolor{CustomLightLightPurple}{HTML}{E6DEFF}
\definecolor{MyPurple}{RGB}{135,58,208}

\newcommand{\modelnamenc}{CoDA\xspace}
\newcommand{\modelnamegradient}{%
  \textbf{%
    \gradientRGB{Co}{238,160,207}{89,190,225}%
    \gradientRGB{DA}{89,190,225}{139,207,79}%
  }\xspace}

\definecolor{CoDABest}{HTML}{EEE7FF}
\definecolor{CoDASecond}{HTML}{DDF7F2}

\newcommand{\best}[1]{\cellcolor{CoDABest}\textbf{#1}}
\newcommand{\second}[1]{\cellcolor{CoDASecond}#1}

\title{\modelnamegradient: \underline{Co}-Evolution of \underline{D}ata Engines and \underline{A}ction Policies}
\runningtitle{CoDA: Co-Evolution of Data Engines and Action Policies}
\newcommand{\affmark}[1]{\textsuperscript{\textcolor{PlanAccent}{#1}}}
\authorblock{\normalfont\fontsize{10}{13}\selectfont
  Tianjiao Yu\affmark{1}\quad
  Zhilin Ou\affmark{2}\quad
  Yifan Shen\affmark{1}\quad
  Xinzhuo Li\affmark{1}\\[3pt]
  Yuanzhe Liu\affmark{1}\quad
  Onkar Kishor Susladkar\affmark{1}\quad
  Fan Wang\affmark{3}\quad
  Ismini Lourentzou\affmark{1}\\[5pt]
  {\fontsize{9}{11}\selectfont
  \textsuperscript{1}University of Illinois Urbana-Champaign \quad 
  \textsuperscript{2}The Chinese University of Hong Kong, Shenzhen\quad
  \textsuperscript{3}Amazon}
}
\authornote{%
  \href{mailto:ty41@illinois.edu}{\texttt{ty41@illinois.edu}}\quad
  \href{mailto:lourent2@illinois.edu}{\texttt{lourent2@illinois.edu}}
}
\newcommand{\institutionlogos}{%
  \includegraphics[height=18pt]{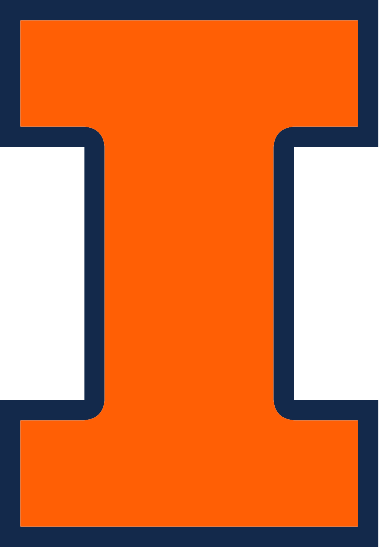}%
  \hspace{17pt}%
  \includegraphics[height=18pt]{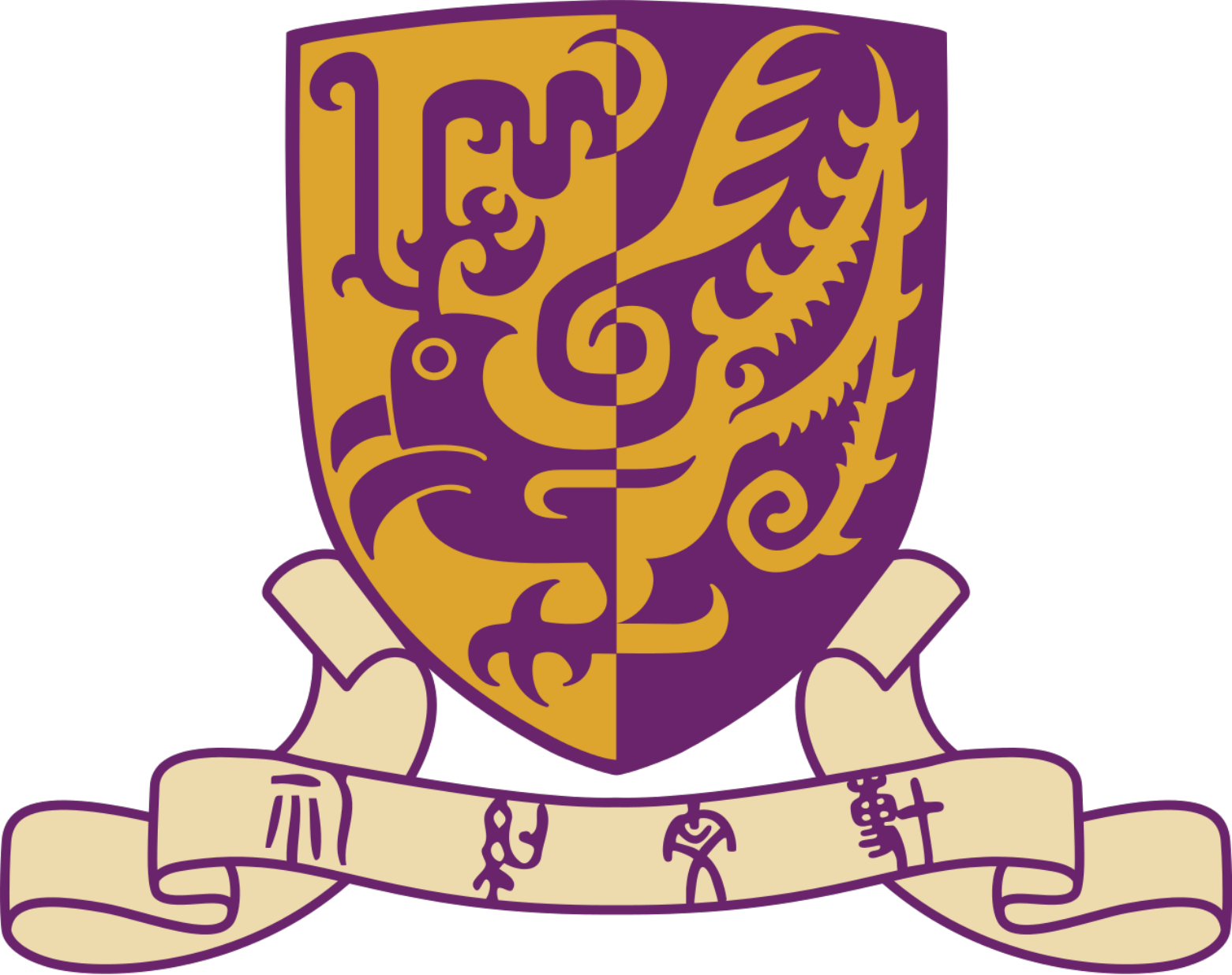}%
  \hspace{17pt}%
  \includegraphics[viewport=15bp 12bp 225bp 228bp,clip,height=18pt]{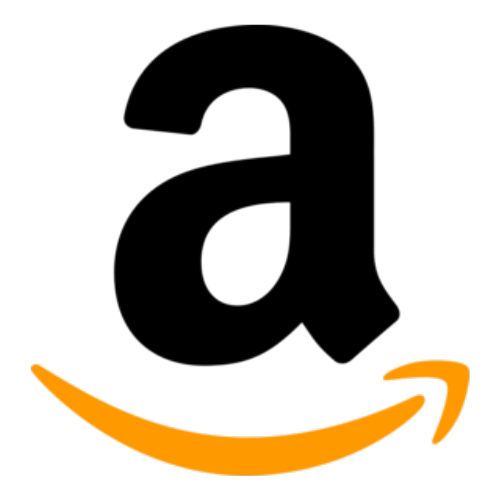}%
  \hfill
  \includegraphics[height=18pt]{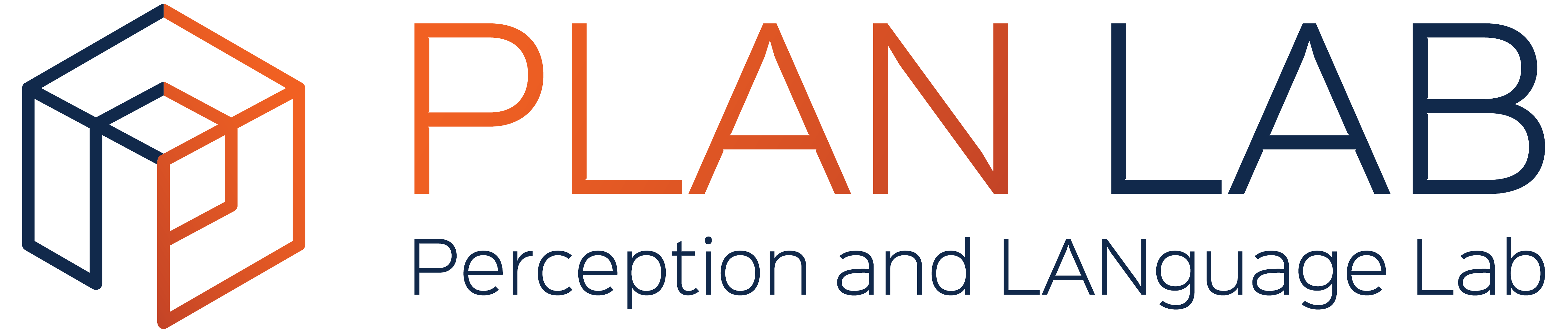}}
\headerlogos{\institutionlogos}
\hypersetup{pdfauthor={Tianjiao Yu, Zhilin Ou, Yifan Shen, Xinzhuo Li, Yuanzhe Liu, Onkar Kishor Susladkar, Fan Wang, Ismini Lourentzou}}
\projecturl{https://plan-lab.github.io/coda}
\paperlinks{\projectlinks}

\teaserposition{afterabstract}
\teaser{%
  \includegraphics[width=\linewidth]{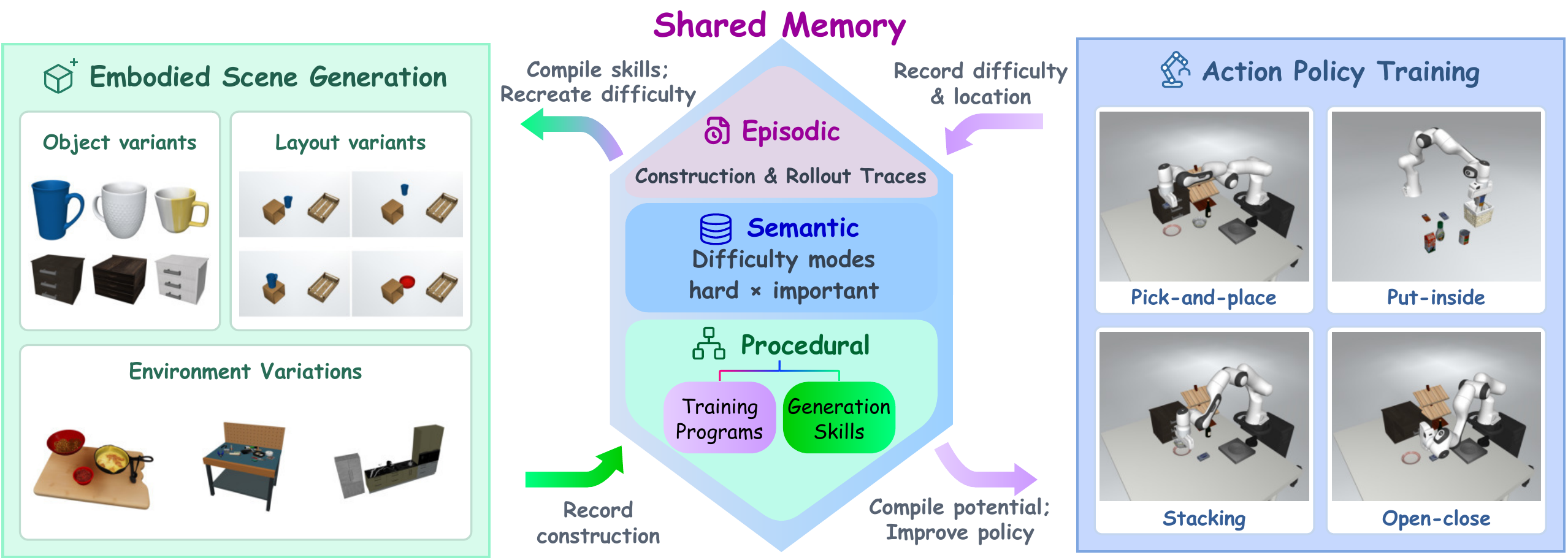}
  \caption{\textbf{\modelnamegradient Co-evolution of data generation and action policies.} Shared memory integrates scene-construction experience and policy rollout diagnoses to guide both what environments to generate and how to train the policy. Generation programs produce environments targeting current policy difficulty modes, while training programs supply shaping potentials for policy improvement. Subsequent rollouts update the memory and guide the next round.}
  \label{fig:teaser}
  \par\vspace{4pt}
  {\color{black!20}\hrule height 0.35pt}
  \vspace{4pt}
}

\definecolor{WhetPink}{RGB}{238,160,207}
\definecolor{WhetBlue}{RGB}{89,190,225}
\definecolor{WhetGreen}{RGB}{139,207,79}
\newcommand{\pmse}[1]{{\scriptsize\,$\pm$#1}}
\providecommand{\captionbox}[2]{}
\renewcommand{\captionbox}[2]{%
  \begingroup
  \setlength{\fboxsep}{1.2pt}%
  \colorbox{#1}{\strut #2}%
  \endgroup
}

\begin{document}
\begin{abstract}
Generative simulation can automate robot environment construction,
but physically valid environments are not necessarily useful
for training.
The challenge is to identify environments that address a policy's
failures and supervision that helps it overcome them.
We introduce \modelnamegradient{}, a framework in which a
simulation data engine and an action policy co-evolve through
a shared heterogeneous memory of environment construction,
policy failures, and learning progress.
The memory compiles validated diagnoses from a vision-language
critic into generation programs that recreate difficulties
and training programs that help the policy overcome them. We further propose \textbf{Text2potential} to convert these diagnoses into staged potentials for action-chunk value learning, providing dense supervision for policy updates while preserving the task's optimal policy under standard shaping assumptions.
Measured learning progress guides further generation and
program refinement.
Across four manipulation task families, \modelnamenc{} improves
out-of-distribution success from 5.3\% after supervised
fine-tuning to 27.6\%, exceeding the strongest closed-loop
baseline by 9.5 percentage points with the same generation budget.
On LIBERO-Pro, 15 rounds of generation and training raise
success from 25\% to 84\% under position perturbations and
from 1\% to 76\% under instruction perturbations.
\end{abstract}
\maketitle
\makeatletter
\setlength{\@fptop}{0pt}
\setlength{\@dblfptop}{0pt}
\setlength{\@fpsep}{16pt}
\setlength{\@dblfpsep}{16pt}
\makeatother
\setlength{\parskip}{3pt plus 1pt minus 1pt}

\section{Introduction}
\label{sec:intro}

Simulation has become a primary source of robot manipulation data, but building simulated
environments has long been the bottleneck: each one requires a task design, semantically
appropriate assets, a plausible layout, and a training signal, often all specified by hand
\citep{wang2023robogen}. Generative simulation removes much of that cost. Coding agents now
synthesize scenes, articulations, and task specifications directly
\citep{wang2023robogen, xia2026sage, yang2026sceneweaver, wang2026embodiedgen,
pfaff2026scenesmith}, and language models author the reward functions that accompany them
\citep{xie2024text2reward, yu2023language}. Environments are no longer scarce.\looseness=-1

A central challenge remaining is deciding which environments to build next and what supervision will help the policy to learn from them.
Existing engines sample variation from hand-specified ranges \citep{ranawaka2026simfoundry}, refine a scene until it is physically valid \citep{xia2026sage, yang2026sceneweaver},
or allocate budget by per-task success rate \citep{kang2026simworld}. 
These signals indicate whether an
environment is valid and how well the current policy performs, but do not
by themselves measure how much the policy improves through training. 
Without such feedback,
a generator risks spending its budget on environments the policy has already mastered or cannot yet learn to solve.\looseness-1

\modelnamegradient closes this loop with a heterogeneous memory shared by the data generator and the action policy. It records repairs needed to make scenes valid, together with policy failure diagnoses grounded in rollout evidence and localized to one generative factor and one trajectory phase. The memory tracks learning progress for recurring difficulty modes to guide subsequent generation.
This evidence is compiled into separate programs for scene generation and policy training, with the latter specifying subgoals and learning tools \citep{wang2023robogen, xie2024text2reward}. 

As tasks involve longer sequences of dependent subtasks, sparse success signals can slow value learning, while designing effective dense rewards can require manual effort. We therefore introduce \textbf{\mempot}, which compiles a failure diagnosis into a potential over the simulator state and uses it to densify value targets for action chunks \citep{li2026reinforcement, giridhar2026beyond}. The resulting value estimates guide policy updates through weighted regression on the policy's own rollouts, favoring action chunks with higher estimated value.\looseness-1

Across four manipulation tasks, \modelnamenc{} achieves 27.6\% out-of-distribution success, compared with 5.3\% for the supervised fine-tuning baseline.
With the same budget for generating environments, a loop guided by per-task success rates achieves 18.1\%, while open-loop variants yield 8.5\%. Extending our loop to more rounds increases out-of-distribution success to 42.1\%. Moreover, on LIBERO-Pro, training on source-task variants generated by \modelnamenc{} increases success from 25\% to 84\% on position generalization and from 1\% to 76\% on task generalization. Finally, we observe that reusing memory across policy backbones transfers accumulated experience, enabling a second backbone to increase out-of-distribution success from 3.8\% to 21.1\%.  

In summary, the contributions of our work are:
\begin{itemize}[itemsep=0ex, parsep=0pt, topsep=-1ex, leftmargin=0.4cm]
\item \textbf{Closed-loop co-evolution.} 
We introduce \modelnamenc{}, a framework that co-evolves a simulation data engine with a trainable action policy. Failure diagnoses grounded in policy rollouts guide environment generation and training, while measured learning progress determines which difficulties to revisit.

\item \textbf{Shared heterogeneous memory.} We introduce a memory shared by the generator and policy that combines episodic evidence, reusable difficulty modes, and procedural programs. The same diagnosis compiles into a generation program that recreates the difficulty and a training program that helps the policy overcome it. Both programs are indexed by difficulty mode, making them reusable across tasks.\looseness-1

  \item \textbf{Dense supervision through \mempot.} Text2potential compiles failure diagnoses into potentials over simulator states that provide dense targets for learning the value of action chunks~\citep{li2026reinforcement}. Reward shaping requires potential evaluations only at chunk endpoints and preserves the policies that maximize the original task return.\looseness-1

  \item \textbf{Improved out-of-distribution generalization.} Across four manipulation task families, \modelnamenc{} achieves 27.6\% out-of-distribution success, compared with 5.3\% after supervised fine-tuning and 18.1\% for success-rate-guided generation using the same accepted-environment budget. On LIBERO-Pro, fifteen rounds of \modelnamenc{} training increase $\pi_{0.5}$'s success from 25\% to 84\% under position perturbations and from 1\% to 76\% under instruction perturbations. Reusing memory across policy backbones also improves OpenVLA-OFT's out-of-distribution success from 3.8\% to 21.1\% without additional environment generation.
\end{itemize}

\section{Related Work}
\label{sec:related}

\textbf{Agentic asset and scene synthesis.}
Recent methods improve visual asset quality~\citep{xiang2025structured,zhao2025hunyuan3d} and model part structure, articulation, or physical properties~\citep{wang2025embodiedgen,le2025articulate,zhou2026articraft,cao2026physx,yu2026part,lu2025dreamart,yu2026dreampartgen,cao2026physxomni}. For robot learning, RoboGen~\citep{wang2023robogen} generates tasks, scenes, and training supervision, while RoboTwin 2.0~\citep{chen2025robotwin} scales bimanual manipulation data through expert data synthesis and domain randomization. SimFoundry~\citep{ranawaka2026simfoundry} generates variations across objects, scenes and tasks, and agentic systems automate scene construction and refinement for simulation~\citep{yang2026sceneweaver,yang2024holodeck,xia2026sage,pfaff2026scenesmith,wang2026embodiedgen}. Our work focuses on adapting environment generation to the policy's learning progress, using feedback from training to guide what is generated next.

\textbf{Curriculum learning and policy improvement.}
Curriculum methods prioritize environments using regret, learning potential, or parameter changes~\citep{dennis2020emergent, parker2022evolving, jiang2021prioritized, yuan2026pace, portelas2020automatic}. POET~\citep{wang2019paired} coevolves agents and parameterized environments, while language models synthesize rewards~\citep{ma2024eureka, xiao2026enpire, xie2024text2reward} and generate adaptive environment programs~\citep{liang2024environment, kang2026simworld}. For policy improvement, VLA policies learn from their rollouts using sparse success rewards~\citep{tan2025interactive, li2026simplevla}. Q-Planning~\citep{giridhar2026beyond} improves action selection while keeping the base policy frozen. We adopt a critic over action chunks~\citep{li2026reinforcement} with advantage-weighted regression to update the policy~\citep{peng2019advantage}. Lifelong methods retain reusable prompts or skill libraries~\citep{yao2025think, lu2026aspire}. \modelnamenc{} links scene generation and policy learning through failure diagnoses localized to scene factors and trajectory phases.\looseness-1
\begin{figure*}[t]
  \centering
  \includegraphics[width=0.98\linewidth]{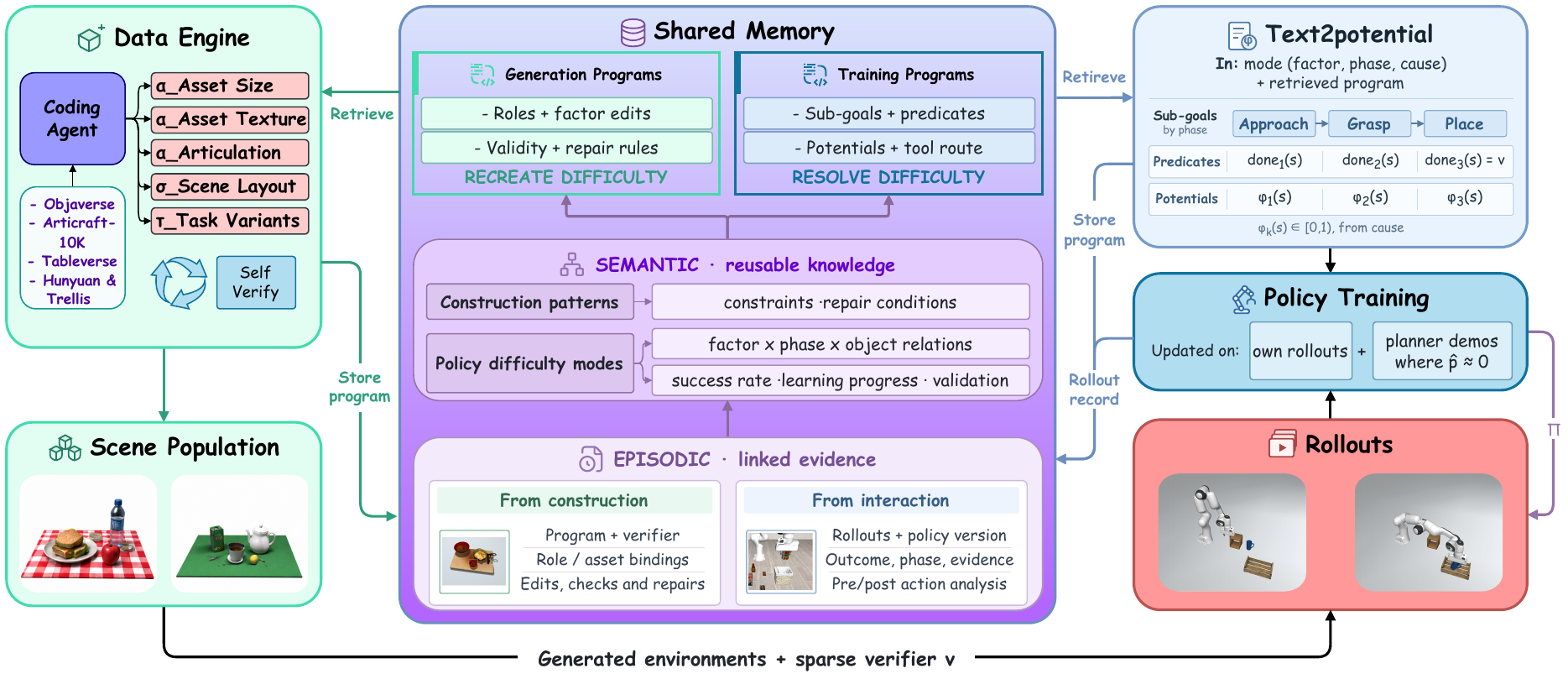}
 \vspace{-0.2cm}
  \caption{\textbf{The \modelnamegradient{} framework.} The data engine  $\eng$ generates executable scenes with independently editable task, spatial, and object-attribute factors $(\tsk,\spa,\att)$ and a sparse success verifier $\vfy$. Construction traces and rollout-grounded failure diagnoses enter a shared memory, which organizes them into reusable difficulty modes. Validated modes are compiled into generation programs that reproduce the diagnosed conditions and training programs that specify subgoals, potentials, and learning tools. \mempot{} supplies dense feedback through chunk-level value targets that guide policy updates, and measured learning progress guides the next generation round.}
  \label{fig:pipeline}

\end{figure*}

\section{Method}
\label{sec:method}
\label{sec:method:overview}

\modelnamegradient{} couples scene generation and policy learning through a shared memory that guides both. A data engine $\eng$, implemented as a tool-augmented coding agent, writes executable scene programs against a simulator SDK. Each program defines an environment and a machine-checkable success predicate $\vfy$. An action policy $\pol$ collects rollouts in these environments, and a vision--language critic $\crit$ diagnoses failures by identifying the responsible factor, a trajectory phase, and supporting rollout evidence. Failures intrinsic to motor execution are recorded separately from those attributed to scene factors. The memory $\mem$ consolidates this evidence into reusable
difficulty modes. Validated modes are compiled into two
complementary outputs: generation programs that recreate the diagnosed conditions and training programs that help the policy $\pol$ overcome them.

We first construct \textbf{seed environments} (\Cref{sec:method:seed}) and define the \textbf{shared memory} (\Cref{sec:method:memory}). Each round then uses the memory to guide  \textbf{scene generation}
(\Cref{sec:method:scene}), compiles validated diagnoses into training programs through \textbf{\mempot} (\Cref{sec:method:t2p}), and updates the policy using \textbf{shaped value estimates} (\Cref{sec:method:train}). The resulting learning progress informs subsequent generation. \Cref{fig:pipeline} provides an overview of the overall framework.

\subsection{Seeding and Scene Representation}
\label{sec:method:seed}
Seeding begins with an object pool~\citep{deitke2023objaverse,wang2026tableverse,zhou2026articraft}. We use a task family's assets when available and otherwise sample from a predefined pool. Given the sampled objects, the engine proposes a task~\citep{wang2023robogen} and specifies its instruction, participating objects, initial and terminal object states, and executable success predicate $\vfy$. Proposals without a verifier are discarded.\looseness-1

Instead of committing to a specific data generation scheme, the engine writes \emph{factored scene programs}. Task, layout, and object attributes can then be perturbed independently, letting the loop test a diagnosis by changing one factor while keeping the others fixed.
Failure diagnosis and scene generation need to share a common representation. The critic should identify factors that the engine can directly modify.
We parameterize each environment as $\env\!=\!(\tsk, \spa, \att)$, generated by executing a scene program $\env\!=\!\Exec(\prog)$ with $\prog \sim \eng(\cdot)$. The \emph{task} factor $\tsk$ specifies the goal predicate, instruction, and initial and terminal object states. The \emph{object-spatial} factor $\spa$ specifies poses, support and containment relations, clearances, and occlusion. The \emph{object self-attribute} factor $\att$ specifies geometry, metric scale, mass, friction, articulation, and appearance.
The SDK exposes these factors through \sdk{set\_task}, \sdk{set\_layout}, and \sdk{set\_attributes}, so the engine can resample any one factor while keeping the other two fixed. This factorization provides a common representation for the
critic's diagnoses, the engine's interventions, and the
applicability conditions of compiled generation programs.

\textbf{Verification.} Each program emits a sparse, machine-checkable success predicate $\vfy$ over its scene objects. This predicate determines task success independently of the critic's judgments and provides the common success criterion across the methods we compare. Dense training feedback is introduced separately through potential-based shaping, rather than replacing the task predicate with a synthesized reward~\citep{xie2024text2reward,yu2023language}. Before policy deployment, each scene must pass verification checks for collision-free placement, static stability during a short passive simulation, and plausible metric scale. Programs that fail these checks are excluded from policy rollouts but retained in memory to inform subsequent scene generation.\looseness-1

\subsection{Shared Heterogeneous Memory}
\label{sec:method:memory}
Memory combines two sources of experience: the engine's construction and repair traces, and the policy's rollout outcomes and failure diagnoses. Following the episodic--semantic--procedural organization of language-agent memory~\citep{sumers2023cognitive}, we retain individual records (\emph{episodic tier}), abstract recurring failure patterns (\emph{semantic tier}), and compile reusable programs for generation and training (\emph{procedural tier}).

\textbf{Episodic tier: construction and rollout evidence.}
The append-only trace store $\buf$ records each scene program $\prog$, its verifier $\vfy$, and summaries of the generation
trace, including violated constraints and verification outcomes.
For deployed environments, it also records the success rate $\psucc(\env)$ over $N$ rollouts and the learning progress $\lv(\env)$ obtained from training~\citep{yuan2026pace}.
The vision--language critic $\crit$ uses scene and rollout evidence to produce a failure diagnosis $\attr\!=\!(\axis,p,c,e)$, where $\axis\!\in\!\{\taskax,\spatax,\attax,\execax\}$ identifies the attributed factor, $p$ the trajectory phase, $c$ a natural-language description of the cause, and $e$ the supporting evidence. These diagnoses are stored alongside the corresponding traces. The additional label $\execax$ covers motor-execution failures.

\textbf{Semantic tier: reusable difficulty modes.}
Episodic records are clustered online into \emph{difficulty modes}
${\clus_j}$ using the factor--phase pair $(\axis,p)$ and
similarity between diagnosed causes.
Each mode captures a recurring failure condition that can arise across tasks and policy backbones, such as ``an occluder adjacent to the target on the approach side''. 
We characterize each mode by two quantities, \emph{hardness} and \emph{importance}.
\emph{Hardness} is measured by the policy's success rate $\psucc_j$, with lower rates indicating greater difficulty.
\emph{Importance} $w_j$ is a running score updated from
mode-level learning progress, measured as the change in
policy success rate on that mode before and after
training~\citep{yuan2026pace}.
These two quantities reflect different properties of a mode.
A mode can remain hard even when training on it yields little learning progress, or become easy after substantial improvement.
Each mode also carries a validation status determined by control batches (\Cref{sec:method:scene}), and
only validated modes are compiled into procedural memory.\looseness-1

\textbf{Procedural tier: generation and training programs.}
The two consumers of memory improve differently. The data engine $\eng$ adapts in context by retrieving and executing programs between rounds, while the action policy $\pol$ learns through gradient updates and acts from observations. 
We therefore compile the same validated mode into two complementary programs:
\begin{itemize}
    \item \emph{Generation programs}, stored in $\skpool$, encode how to reproduce and vary the diagnosed scene conditions. 
They retain the parent scene program, validated variants along one factor, compilation-check outcomes, and the required relational structure expressed over symbolic object roles.
\item \emph{Training programs}, stored in $\dirpool$, specify how to address the policy failure through subgoal decompositions, potentials, and learning tools. They retain the diagnosed phase and evidence of which training programs improved policy performance.
\end{itemize}
The generation compiler $\Rgrind$ produces the former, while Text2potential implements the training compiler $\Rhone$ (\Cref{sec:method:t2p}). During training, these programs evaluate subgoal completion and compute potentials that shape value targets. At deployment, the trained policy predicts actions from observations and the task instruction without executing these programs. 

\subsection{Memory-Guided Scene Generation}
\label{sec:method:scene}
\label{sec:method:scene:synth}
\label{sec:method:skillevo}
At each round, memory guides which difficulty modes to revisit and how to instantiate them in new environments. We allocate the generation budget across difficulty modes using an upper-confidence rule based on their hardness and observed learning progress. An $\varepsilon$-fraction of the budget is reserved for novel environments, while a separate replay fraction revisits previously encountered modes to detect performance regressions. 
A validated generation program is added to the pool $\skpool$ when the policy's success rate on its variants falls within the designated success band and is removed when success rate is outside that band. For each selected mode, the engine retrieves a generation program specifying the scene conditions to reproduce and the factor to vary. 
Generation programs express scene conditions in terms of symbolic object roles, allowing the engine to reuse them across task families by assigning those roles to objects in a new scene (Appendix~\ref{app:lifting}).\looseness=-1

For each selected candidate difficulty mode linked to a scene factor, the engine generates a targeted batch that varies the identified factor and a control batch that varies a different factor. After verification checks, we evaluate the current policy on both batches.
If the targeted variants yield a lower policy success rate than the control variants, the proposed difficulty mode is compiled into reusable generation and training programs.
The policy learns from its rollouts using the success predicate $\vfy$ and the potential-based feedback supplied by Text2potential (\Cref{sec:method:t2p}). A scripted planner also attempts each valid variant, and trajectories that satisfy $\vfy$ are retained as demonstrations $\dem_t$, providing additional imitation supervision when available.\looseness-1

\subsection{Text2potential}
\label{sec:method:t2p}
\label{sec:method:train:compile}
The success predicate $\vfy$ determines whether a generated task is complete,
but provides little feedback on intermediate progress. This makes value
learning difficult when success requires a long sequence of actions.
\mempot{} converts a validated failure diagnosis into subgoals and potentials
over simulator states. These potentials provide dense feedback for value learning (\Cref{sec:method:train}). 

Prior language-based reward synthesis provides dense training feedback, but its iterative refinement relies on humans to describe policy failures observed in rollouts, and can produce rewards that favor behavior other than task completion \citep{xie2024text2reward}.
Text2potential instead uses a critic to diagnose failures from the policy’s own rollouts and compiles these diagnoses into a potential function that shapes the value critic’s training targets, providing dense feedback without replacing the original task reward.

Let $\clus_j$ be a validated difficulty mode with record
$\attr=(\axis,p,c,e)$, let $\prog$ be a scene program in $\clus_j$, and let
$\dirpool_j$ contain the training programs previously compiled for that mode,
if any. Text2potential is the map\looseness-1
\begin{equation}
  \begin{aligned}
  &\Rhone : (\attr, \prog, \dirpool_j) \;\longmapsto\\
  &\quad\big(\subg_{1:m},\; \mathrm{done}_{1:m},\; \potsub{1:m},\; \mathrm{tool}_{1:m}\big),
  \end{aligned}
  \label{eq:t2p}
\end{equation}
where $m$ is the number of subgoals and $\subg_{1:m}$ is the sequence obtained
by splitting the task at the diagnosed phase $p$. For each subgoal $k$,
$\mathrm{done}_k\!:\!\mc{S}\to\{0,1\}$ is its completion predicate,
$\potsub{k}\!:\!\mc{S}\to[0,1)$ measures progress within the subgoal, and
$\mathrm{tool}_k$ specifies the learning procedure. Here, $\mc{S}$ is the
simulator state space. The final completion predicate is the task verifier,
$\mathrm{done}_m\equiv\vfy$. The completion predicates and potentials are implemented as functions over the objects and simulator quantities exposed by $\prog$. The subgoal potentials are combined into a
\emph{staged potential}:
\begin{equation}
  \begin{aligned}
  k^{*}(s) &= \sum_{k=1}^{m}\mathrm{done}_k(s),\\
  \pot(s) &=
  \begin{cases}
    k^{*}(s)+\potsub{k^{*}(s)+1}(s), & k^{*}(s)<m,\\
    m, & k^{*}(s)=m.
  \end{cases}
  \end{aligned}
  \label{eq:staged-potential}
\end{equation}
Here, $k^{*}(s)$ counts the subgoal predicates satisfied at state $s$. The compiler is instructed to make completion predicates cumulative, so earlier subgoals remain satisfied as later ones are pursued.
The diagnosed phase $p$ identifies where the policy fails and guides the
task decomposition, while the diagnosed cause $c$ guides the construction
of the subgoal potentials. The scene program provides the object handles,
poses, and relations needed to express subgoal completion predicates and
potentials as functions of simulator state.\looseness=-1

\textbf{Compilation and execution checks.}
The compiler $\Rhone$ uses a language model supplied with the failure record,
the scene program's symbol table, geometric and contact helper functions, and
training programs retrieved from $\dirpool_j$ as examples. 
It produces the
subgoal decomposition, completion predicates, potentials, and learning-tool
assignments in Eq. (\ref{eq:t2p}). 
Each emitted program undergoes syntax checks and, when rollout
states are supplied, runtime checks on those states. Detected
errors are returned to the compiler for repair.
These checks establish that the programs execute. Their usefulness is
assessed through subsequent policy learning
(\Cref{sec:method:train:refine}).\looseness-1

\subsection{Policy Training}
\label{sec:method:train}
The compiled potential shapes the training targets of a value critic
over action chunks. The value critic guides data collection and supplies
the advantage estimates used to weight policy updates. This value
critic is distinct from the vision--language critic $\crit$, which diagnoses
failures. The policy must be able to sample $j$ candidate action chunks conditioned on observations and the task, $(o,\tsk)$, and support a weighted supervised action loss. \looseness=-1

 \textbf{Chunk-level value learning.}
\label{sec:method:train:value}
The policy takes observations and the task instruction as input
and emits action chunks $\chunk{t}$ of length $h$, which are
executed open-loop. The value critic $\Qf(o_t,\chunk{t})$ takes
observations, the task instruction, and an action chunk as input,
using visual and language encoders separate from the policy's.
We train a value critic $\Qf(o_t,\chunk{t})$ with an $h$-step
temporal-difference target~\citep{li2026reinforcement}. For sparse task reward
$r_t$ supplied by $\vfy$ and discount factor $\gamma$, potential-based shaping
defines $\shaped_t=r_t+\beta\big[\gamma\pot(s_{t+1})-\pot(s_t)\big]$ for nonterminal transitions, where $s_t$ is the simulator state. Summing over a chunk of length $h$ gives
\begin{equation}
  \begin{aligned}
  \sum_{i=0}^{h-1} \gamma^{i} \shaped_{t+i}
  &= \underbrace{\sum_{i=0}^{h-1} \gamma^{i} r_{t+i}}_{\text{sparse return}} + \beta\gamma^{h}\pot(s_{t+h}) - \beta\pot(s_t),
  \end{aligned}
  \label{eq:telescope}
\end{equation}
so the shaping contribution requires potential evaluations only at the
chunk's endpoints. Following Giridhar et al.~\citep{giridhar2026beyond}, we use the next
action chunk stored in the replay buffer for bootstrapping. For a nonterminal chunk of length $h$, the target using the target critic
$\bar{Q}_{\phi}$ is
\begin{equation}
  \begin{aligned}
  y_t &= \sum_{i=0}^{h-1} \gamma^{i} r_{t+i}
         - \beta\pot(s_t)\\
      &\quad + \gamma^{h}\Big[
         \beta\pot(s_{t+h})
         + \bar{Q}_{\phi}(o_{t+h}, \chunk{t+h})\Big].
  \end{aligned}
  \label{eq:target}
\end{equation}
Both the endpoint-potential term and the critic bootstrap are
omitted at successful task completion, and both successful and
failed rollouts contribute to critic training. Failed rollouts
have zero sparse task reward, but can have nonzero shaped targets.
The value critic is trained by categorical value
regression~\citep{giridhar2026beyond}, with value bins spanning
$[-\beta m_{\max},1]$, where $m_{\max}$ is the maximum number
of subgoals.\looseness-1

\begin{table*}[t]
  \centering\small\setlength{\tabcolsep}{3.5pt}
  \caption{\textbf{Policy generalization across four manipulation task families.} OOD success rates (\%) under different generation and training strategies. \textbf{In} denotes success on each generator's own distribution. Values are mean $\pm$ standard error over three seeds. \protect\captionbox{CoDABest}{\textbf{Best}} and \protect\captionbox{CoDASecond}{second best} highlighted.}

  \label{tab:main}
    
  \vspace{-0.2cm}
  
  \begin{tabularx}{\linewidth}{@{}>{\raggedright\arraybackslash}Xcccccc@{}}
    \toprule
    & \multicolumn{4}{c}{\textbf{OOD by family}} & \multicolumn{2}{c}{\textbf{Average}} \\
    \cmidrule(lr){2-5}\cmidrule(lr){6-7}
    \textbf{Data generation} & Pick-place & Articulated & Put-inside & Stack/insert & In & OOD \\
    \midrule
    pretrained, zero-shot                                    & 3.1\pmse{0.6} & 1.6\pmse{0.5} & 2.7\pmse{0.6} & 1.8\pmse{0.6} & 3.4\pmse{0.2} & 2.3\pmse{0.3} \\
    + seed demonstrations (SFT)                              & 6.4\pmse{0.6} & 4.7\pmse{0.4} & 5.9\pmse{0.5} & 4.2\pmse{0.4} & 11.7\pmse{0.4} & 5.3\pmse{0.3} \\
    domain randomization                                     & 8.8\pmse{0.4} & 8.3\pmse{0.6} & 7.6\pmse{0.4} & 6.9\pmse{0.5} & 17.2\pmse{0.3} & 7.9\pmse{0.4} \\
    uniform variants \citep{ranawaka2026simfoundry}          & 10.6\pmse{0.4} & 6.9\pmse{0.6} & 9.7\pmse{0.4} & 6.8\pmse{0.3} & 21.6\pmse{0.3} & 8.5\pmse{0.2} \\
    scalar-feedback co-evolution \citep{kang2026simworld}    & 20.9\pmse{0.7} & 15.3\pmse{0.7} & \second{19.6\pmse{1.2}} & 16.6\pmse{0.7} & \second{33.8\pmse{0.4}} & 18.1\pmse{0.5} \\
    our critic, no memory                                    & \second{22.4\pmse{0.7}} & \second{17.8\pmse{0.9}} & 18.3\pmse{1.3} & \second{18.3\pmse{0.8}} & 31.7\pmse{0.6} & \second{19.2\pmse{0.7}} \\
    \midrule
    \rowcolor{gray!8}
    \modelnamegradient{} (ours)                              & \best{33.1\pmse{1.2}} & \best{24.7\pmse{0.9}} & \best{28.9\pmse{1.3}} & \best{23.7\pmse{1.0}} & \best{39.4\pmse{0.5}} & \best{27.6\pmse{0.5}} \\
    \bottomrule
  \end{tabularx}
  
\end{table*}

\textbf{Policy update.}
\label{sec:method:train:act}
During data collection, we sample $j$ candidate chunks from the current policy
and score each chunk with $\Qf$. We apply a softmax to
these scores and execute the weighted average of the candidate chunks.
The resulting rollouts provide training data for the value critic and the policy. After fitting
the critic based on Eq.~(\ref{eq:target}), we update the policy by advantage-weighted
regression~\citep{peng2019advantage}:\looseness-1
\begin{equation}
  \begin{aligned}
  \loss_{\pi}&=\E\Big[\, e^{A(o,\chunk{t})/\lambda}\; \ell_{\mathrm{BC}}\big(\pol;\, o, \chunk{t}\big) \Big],\\
  A(o, a)&=\Qf(o, a) - \tfrac{1}{j}\textstyle\sum_{a' \sim \pol(\cdot \mid o)} \Qf(o, a'),
  \end{aligned}
  \label{eq:awr}
\end{equation}
where $\lambda>0$ controls the advantage weighting and
$\ell_{\mathrm{BC}}$ is the policy's supervised action loss. Task conditioning is omitted from the notation for readability. The baseline
averages the critic's values over $j$ chunks sampled from the policy at the
same observation. The advantage weight scales the action loss. Computing this weight  does not require an action-density evaluation. The same weighting can
therefore be applied to the supervised losses of discrete,
flow-matching, and diffusion policies~\citep{li2026reinforcement}.
Planner demonstrations provide additional imitation supervision
and contribute to the supervised loss with unit weight. At deployment, the trained policy predicts action chunks directly, without critic-guided candidate averaging.

For a fixed, bounded potential with appropriate terminal-state handling,
potential-based shaping preserves the policies that maximize the original
task return~\citep{ng1999policy}. For any fixed policy, exact action and
state values shift by the same term, $-\beta\pot(s)$, leaving action advantages
unchanged:
$A_{\pot}=(Q-\beta\pot)-(V-\beta\pot)=A$.
Under these conditions, shaping preserves the policies that maximize the
original task return. This property concerns exact values; the effect of shaping on the learned critic and policy is assessed through subsequent policy progress.\looseness=-1

\begin{figure*}[t!]
  \centering
  \resizebox{\linewidth}{!}{\begin{tikzpicture}
\pgfplotsset{
  every axis/.append style={
    scale only axis, width=3.25cm, height=2.9cm,
    tick label style={font=\scriptsize}, label style={font=\scriptsize},
    title style={font=\scriptsize\bfseries, yshift=-0ex},
    axis line style={gray!50}, tick style={gray!50},
    ymajorgrids, grid style={gray!15},
    legend style={font=\scriptsize, draw=none, fill=none, inner sep=1pt, row sep=-3pt},
  },
  /tikz/lift/.style={font=\tiny, text=gray!60!black, anchor=south, inner sep=0.5pt},
  bar legend/.style={legend image code/.code={\fill[##1] (0cm,-0.09cm) rectangle (0.24cm,0.09cm);}},
}
\begin{groupplot}[group style={group size=3 by 1, horizontal sep=2cm}]

\nextgroupplot[
  title={(a) OOD success vs.\ budget},
  xlabel={accepted environments per family}, ylabel={OOD success (\%)},
  xmin=0, xmax=960, ymin=0, ymax=50, xtick={0,320,640,960}, ytick={0,20,40},
  legend to name=leg:headline, legend columns=3,
  legend style={/tikz/every even column/.append style={column sep=0.5em}},
]
\addplot[name path=seup, draw=none, forget plot] coordinates {(0,5.9) (64,12.1) (128,17.2) (192,21.5) (256,25.3) (320,28.1) (384,31) (448,33.8) (512,35.6) (576,37.4) (640,39.5) (704,41.5) (768,42.3) (832,42.7) (896,43) (960,43.4)};
\addplot[name path=selo, draw=none, forget plot] coordinates {(0,4.7) (64,10.7) (128,15.6) (192,20.1) (256,23.5) (320,27.1) (384,29) (448,31.6) (512,33.6) (576,35) (640,37.3) (704,38.9) (768,39.9) (832,40.1) (896,40.2) (960,40.8)};
\addplot[WhetBlue!35, forget plot] fill between[of=seup and selo];
\addplot[WhetBlue!80!black, very thick] coordinates {(0,5.3) (64,11.4) (128,16.4) (192,20.8) (256,24.4) (320,27.6) (384,30) (448,32.7) (512,34.6) (576,36.2) (640,38.4) (704,40.2) (768,41.1) (832,41.4) (896,41.6) (960,42.1)};
\addlegendentry{\modelnamenc{} (ours)}
\addplot[WhetBlue!80!black, thick, dashed] coordinates {(0,5.3) (64,9.9) (128,13.5) (192,16.4) (256,18.1) (320,19.2) (384,21.9) (448,21.3) (512,22.3) (576,22.8) (640,23.6) (704,24) (768,18.6) (832,20.8) (896,25.2) (960,25.3)};
\addlegendentry{our critic, no memory}
\addplot[WhetPink!80!black, thick] coordinates {(0,5.3) (64,9.7) (128,12.8) (192,15.4) (256,16.9) (320,18.1) (384,19) (448,19.3) (512,20.1) (576,19.9) (640,20.4) (704,20.8) (768,20.8) (832,21.1) (896,21) (960,21.2)};
\addlegendentry{scalar feedback}
\addplot[gray!70!black, thick] coordinates {(0,5.3) (64,6.6) (128,7.6) (192,8.5) (256,9) (320,8.5) (384,8.8) (448,9.3) (512,9.5) (576,10.9) (640,10) (704,10.2) (768,10.2) (832,10.4) (896,10.5) (960,11.5)};
\addlegendentry{uniform variants}
\addplot[gray!70!black, thick, dotted] coordinates {(0,5.3) (64,6.2) (128,6.8) (192,7.3) (256,7.65) (320,7.9) (384,8.2) (448,8.4) (512,8.7) (576,8.8) (640,9) (704,9.1) (768,9.1) (832,9.3) (896,9.3) (960,9.4)};
\addlegendentry{domain randomization}
\addplot[gray!50, thick] coordinates {(0,5.3) (960,5.3)};
\addlegendentry{SFT only}
\draw[gray!60, dashed] (axis cs:320,0) -- (axis cs:320,50);
\node[font=\tiny, text=gray!60!black, anchor=south west, inner sep=1pt] at (axis cs:325,45) {\Cref{tab:main}};

\nextgroupplot[
  title={(b) difficulty modes over rounds},
  xlabel={round}, ylabel={difficulty modes},
  ybar stacked, bar width=3.4pt, bar legend,
  xmin=0.3, xmax=15.7, ymin=0, ymax=42, xtick={1,5,10,15}, ytick={0,10,20,30},
  reverse legend, legend columns=2, 
  legend style={at={(0.5,0.98)}, anchor=north, /tikz/every even column/.append style={column sep=0.4em}},
]
\addplot[fill=WhetGreen, draw=none] coordinates {(1,0) (2,2) (3,6) (4,5) (5,8) (6,4) (7,7) (8,5) (9,3) (10,6) (11,4) (12,5) (13,3) (14,4) (15,1)};
\addlegendentry{leaving}
\addplot[fill=WhetBlue, draw=none] coordinates {(1,0) (2,10) (3,13) (4,15) (5,13) (6,15) (7,13) (8,13) (9,14) (10,12) (11,13) (12,11) (13,9) (14,5) (15,4)};
\addlegendentry{staying}
\addplot[fill=WhetPink, draw=none] coordinates {(1,12) (2,9) (3,7) (4,6) (5,6) (6,5) (7,5) (8,4) (9,4) (10,5) (11,3) (12,1) (13,0) (14,0) (15,1)};
\addlegendentry{entering}
\node[lift] at (axis cs:1,12) {0};
\node[lift] at (axis cs:2,21) {2};
\node[lift] at (axis cs:3,26) {3};
\node[lift] at (axis cs:4,26) {4};
\node[lift] at (axis cs:5,27) {3};
\node[lift] at (axis cs:6,24) {4};
\node[lift] at (axis cs:7,25) {2};
\node[lift] at (axis cs:8,22) {3};
\node[lift] at (axis cs:9,21) {3};
\node[lift] at (axis cs:10,23) {5};
\node[lift] at (axis cs:11,20) {2};
\node[lift] at (axis cs:12,17) {1};
\node[lift] at (axis cs:13,12) {0};
\node[lift] at (axis cs:14,9) {0};
\node[lift] at (axis cs:15,6) {1};
\nextgroupplot[
  title={(c) OOD success improvement per round},
  xlabel={round}, ylabel={OOD success gained (\% pts)},
  ybar, bar width=2.6pt, bar legend, tick pos=left,
tick align=inside,
  xmin=0.3, xmax=15.7, ymin=-0, ymax=6.5, xtick={1,5,10,15}, ytick={0,2,4,6}, 
  legend style={at={(0.97,0.97)}, anchor=north east},
]
\addplot[fill=WhetBlue!80!black, draw=none] coordinates {(1,6.1) (2,5) (3,4.4) (4,3.6) (5,3.2) (6,2.4) (7,2.7) (8,1.9) (9,1.6) (10,2.2) (11,1.8) (12,0.9) (13,0.3) (14,0.2) (15,0.5)};
\addlegendentry{\modelnamenc{} (ours)}
\addplot[fill=WhetPink!80!black, draw=none] coordinates {(1,4.4) (2,3.1) (3,2.6) (4,1.5) (5,1.2) (6,0.9) (7,0.3) (8,0.8) (9,-0.2) (10,0.5) (11,0.4) (12,0) (13,0.3) (14,-0.1) (15,0.2)};
\addlegendentry{task feedback}
\draw[gray!60, dashed] (axis cs:5.5,0) -- (axis cs:5.5,6.5);

\end{groupplot}
\path (group c1r1.south west) -- (group c3r1.south east) node[midway, below=0.75cm] {\pgfplotslegendfromname{leg:headline}};
\end{tikzpicture}}
\vspace{-0.8cm}
  \caption{\textbf{Co-evolution over rounds.} (a) Average OOD success as number of accepted environments per task family increases. Dashed line marks the comparison reported in \Cref{tab:main}. Curves show mean over three seeds, with standard error shaded for \modelnamenc{}.
  (b) Number of difficulty modes newly selected for training, retained for further training, and removed from the memory in each round. Labels above bars separately report the number of generation programs reused from other task families.
  (c) OOD success gain (percentage points) per round by \modelnamenc{} and generation guided by task success rates.}
  \label{fig:headline}

\end{figure*}
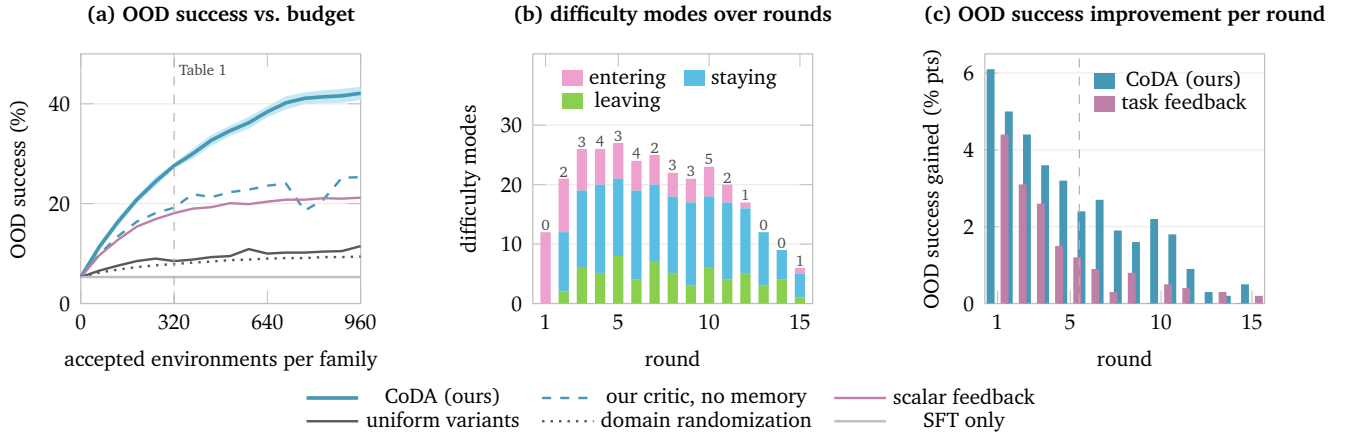

\textbf{Refinement and routing.}
\label{sec:method:train:refine}
\label{sec:method:train:route}
We assess each training program by measuring the policy's improvement on the corresponding difficulty mode, using $\vfy$ to determine task success. We also compare how quickly the value critic converges with and without the compiled potential. If the potential provides no improvement in convergence speed over sparse task rewards alone, the vision--language critic $\crit$ revisits the failure diagnosis, and $\Rhone$ revises the subgoal decomposition.
For each subgoal, we select a learning procedure based on the policy's measured success rate~\citep{wang2023robogen}. When this rate is near zero and a planner demonstration is available, we use behavior cloning on $\dem_t$ to teach the policy an initial solution. When the success rate falls within the target band, we use Text2potential to shape the value-learning targets. For difficulty modes with success rates above the band, we periodically re-evaluate the policy to detect performance regressions. If none of the subgoals falls within the target band, we return the difficulty mode to the data engine for further environment generation and reduce its importance score when allocating subsequent generation budgets.\looseness-1

\textbf{Scheduling.}
Each round generates $B$ targeted environments and their controls.
After verification, we collect $N$ policy rollouts in each valid environment.
The vision--language critic examines a stratified
sample of at most $C$ failed rollouts per mode. Between generation rounds,
the value critic receives $K_Q$ updates and the policy receives $K_\pi$
updates using the accumulated buffer. After validation, difficulty modes are compiled into training programs for policy learning and generation programs for subsequent rounds.
A new generation round is triggered when fewer than eight difficulty modes remain within the success band or when learning progress plateaus for 2k policy-update steps.
The loop terminates when no mode remains in the band or the generation budget
is exhausted.\looseness-1

\begin{table*}[t!]
  \centering\small\setlength{\tabcolsep}{3pt}
  \caption{\textbf{LIBERO-Pro generalization to position and instruction perturbations.} Success rates under position (Pos) and instruction (Task) perturbations for the Object, Goal, and Spatial suites. StdLBR reports success on standard LIBERO. \\\protect\captionbox{CoDABest}{\textbf{Best}} and \protect\captionbox{CoDASecond}{second best} highlighted.}
  \label{tab:liberopro}
    \vspace{-0.2cm}
  \begin{tabularx}{\linewidth}{@{}>{\raggedright\arraybackslash}Xc|cccccc|cc@{}}
    \toprule
    & & \multicolumn{2}{c}{\textbf{Object}} & \multicolumn{2}{c}{\textbf{Goal}} & \multicolumn{2}{c|}{\textbf{Spatial}} & \multicolumn{2}{c}{\textbf{Overall}} \\
    \cmidrule(lr){3-4}\cmidrule(lr){5-6}\cmidrule(lr){7-8}\cmidrule(lr){9-10}
    \textbf{Method} & \textbf{StdLBR} & Pos & Task & Pos & Task & Pos & Task & Pos & Task \\
    \midrule
    OpenVLA \citep{kim2024openvla}          & 0.765 & 0.00 & 0.00 & 0.00 & 0.00 & 0.00 & 0.00 & 0.00 & 0.00 \\
    $\pi_0$ \citep{black2024pi_0}                & 0.942 & 0.00 & 0.00 & 0.00 & 0.00 & 0.00 & 0.00 & 0.00 & 0.00 \\
    $\pi_{0.5}$ \citep{intelligence2025pi_}              & 0.971 & 0.17 & 0.01 & 0.38 & 0.00 & 0.20 & 0.01 & 0.25 & 0.01 \\
    Human-written programs                       & ---   & 0.14 & 0.00 & 0.00 & 0.00 & 0.23 & 0.05 & 0.12 & 0.02 \\
    ASPIRE (code-as-policy) \citep{lu2026aspire} & ---   & \best{0.98} & \best{0.95} & \second{0.81} & 0.45 & 0.51 & 0.60 & 0.77 & 0.67 \\
    \midrule
    \modelnamegradient{} ($\pi_{0.5}$), $n$ = 5 rounds     & \second{0.971} & 0.90 & 0.86 & 0.80 & \second{0.58} & \second{0.76} & \second{0.72} & \second{0.82} & \second{0.72} \\
    \rowcolor{gray!8}
    \modelnamegradient{} ($\pi_{0.5}$), $n$ = 15 rounds   & \best{0.974} & \second{0.91} & \second{0.88} & \best{0.83} & \best{0.64} & \best{0.78} & \best{0.76} & \best{0.84} & \best{0.76} \\
    \bottomrule
  \end{tabularx}

\end{table*}

\section{Experiments}
\label{sec:exp}

\begin{table*}[t]
  \centering
  \small

  \setlength{\tabcolsep}{4pt}
  \caption{\textbf{Critic ablations.}
  OOD success at a fixed training budget, diagnoses refuted
  by control evaluations, and agreement with two human annotators.
  \protect\captionbox{CoDABest}{\textbf{Best}} and
  \protect\captionbox{CoDASecond}{second best} results are highlighted.}
  \label{tab:critic}
 \vspace{-0.2cm}
  \begin{tabularx}{\linewidth}{@{}>{\raggedright\arraybackslash}Xccc@{}}
    \toprule
    \textbf{Critic}
    & \shortstack{\textbf{OOD success}\\(\%) $\uparrow$}
    & \shortstack{\textbf{Refuted}\\(\%) $\downarrow$}
    & \shortstack{\textbf{Human agr.}\\(\%) $\uparrow$} \\
    \midrule
    Qwen3-VL-235B (open weights)
      & 24.9\pmse{0.8} & 27.5 & 73.8 \\
    InternVL3-78B (open weights)
      & 25.7\pmse{0.7} & 24.1 & 76.3 \\
    Claude Opus 5
      & \second{26.1\pmse{0.6}}
      & \second{20.3}
      & \second{81.5} \\
    \midrule
    \rowcolor{gray!8}
    GPT-5.5 (default)
      & \best{27.6\pmse{0.5}}
      & \best{18.2}
      & \best{82.4} \\
    \bottomrule
  \end{tabularx}

\end{table*}

\begin{table*}[!t]
  \centering\small\setlength{\tabcolsep}{4pt}
  \caption{\textbf{\mempot{} ablations.}  
We report OOD success (\%) at a fixed training budget,
policy updates needed to reach 25\% OOD success (in thousands),
and agreement (\%) with the sparse-only critic's action-chunk
selections after convergence. \\ \protect\captionbox{CoDABest}{\textbf{Best}} and \protect\captionbox{CoDASecond}{second best} highlighted.}
  \label{tab:t2p}
      
   \vspace{-0.2cm}
      
  \begin{tabularx}{\linewidth}{@{}>{\raggedright\arraybackslash}Xccc@{}}
    \toprule
    \textbf{Training signal} & \textbf{OOD success} & \textbf{Steps to 25\% OOD (k)} & \textbf{Agreement} \\
    \midrule
    sparse verifier only ($\pot \equiv 0$)                                 & 22.3\pmse{0.8} & 31 & --- \\
    potential from task description \citep{xie2024text2reward}             & 24.6\pmse{0.7} & 29 & 95.3 \\
    free-form dense reward, not a potential                                & 23.5\pmse{1.0} & 33 & 72.4 \\
    frozen actor: $\Qf$-weighted selection only \citep{giridhar2026beyond} & 21.4\pmse{0.9} & --- & \second{96.8} \\
    \mempot{} without refinement                                           & \second{26.1\pmse{0.6}} & \second{27} & 95.9 \\
    \midrule
    \rowcolor{gray!8}
    \textbf{\mempot{} (ours)}                                              & \best{27.6\pmse{0.5}} & \best{21} & \best{97.2} \\
    \bottomrule
  \end{tabularx}

\end{table*}

\textbf{Experimental Setup.}
\label{sec:exp:baselines}
\label{sec:exp:metrics}
We evaluate \modelnamenc{} on four manipulation task families and LIBERO-Pro~\cite{zhou2025libero} using $\pi_{0.5}$~\cite{intelligence2025pi_} as the primary policy backbone. Our main comparison includes the released $\pi_{0.5}$ checkpoint, supervised fine-tuning on seed demonstrations, and policies trained with domain randomization, uniform environment variants \citep{ranawaka2026simfoundry}, or success-rate-guided generation \citep{kang2026simworld}. Generation baselines use the same SDK and budgets. We ablate persistent memory and the separation of generation and training stores, comparing against single-store alternatives that include a skill library \citep{wang2023voyager}. We report per-suite success rates and average success across held-out suites, including generalization to unseen object instances, layouts, tasks, and categories. Additional metrics assess generation quality through verification pass rate and coverage entropy, and diagnosis quality through targeting precision, refutation rate, and agreement with two annotators on 100 sampled diagnoses. We report computational cost in LLM tokens, simulation hours, and GPU-hours per 1,000 accepted environments.
Full experimental details are provided in Appendix~\ref{app:impl}.\looseness-1

\textbf{Main results.}
\label{sec:exp:main}
\Cref{tab:main} shows \modelnamenc{} achieves the highest OOD success across all four task families. Average OOD success reaches 27.6\%, compared with 18.1\% for generation guided by task success rates and 19.2\% for our critic without memory, improvements of 9.5 and 8.4 percentage points, respectively. Both substantially exceed uniform generation (8.5\%) and domain randomization (7.9\%), but retaining and reusing experience through our memory yields the strongest performance.
\Cref{fig:headline} examines how performance changes as generation and training continue. Extending \modelnamenc{} from five to fifteen rounds increases OOD success from 27.6\% to 42.1\%, while uniform generation, domain randomization, and success-rate-guided generation yield much smaller additional gains. In later rounds, fewer difficulty modes remain unresolved, and improvements become smaller. Removing memory produces less consistent progress, including two temporary drops in success.\looseness=-1

\textbf{LIBERO-Pro.}
Training on generated variants of the LIBERO-Pro source tasks substantially improves robustness to position and instruction perturbations  (\Cref{tab:liberopro}). After five rounds, \modelnamenc{} achieves overall success rates of 82\% and 72\% under the two perturbations, respectively. Extending training to fifteen rounds raises these rates to 84\% and 76\%, compared with 25\% and 1\% for the initial $\pi_{0.5}$ checkpoint. The fifteen-round model also exceeds ASPIRE~\citep{lu2026aspire} by 7 and 9 percentage points overall, achieving the highest success on the Goal and Spatial suites under both perturbations, while ASPIRE remains strongest on the Object suite. 
These robustness gains preserve performance on standard LIBERO, where success reaches 97.4\%, compared with 97.1\% before adaptation.\looseness=-1

\textbf{Sensitivity to the failure diagnosis critic.}
\Cref{tab:critic} varies the vision--language model used to diagnose failures while keeping the engine and compiler fixed. Replacing GPT-5.5 with Claude Opus 5, InternVL3-78B, or Qwen3-VL-235B slightly reduces OOD success from 27.6\% to 26.1\%, 25.7\%, and 24.9\%, respectively. Models with lower agreement with human annotators also produce more 
diagnoses rejected by the control tests. Nevertheless, all four models outperform generation guided by task success rates: even Qwen3-VL-235B exceeds that baseline by 6.8 percentage points. The framework therefore remains effective across the tested diagnosis models.\looseness-1

\begin{figure*}[!t]
  \centering
  \vspace{-0.2cm}
  \includegraphics[height=9cm]{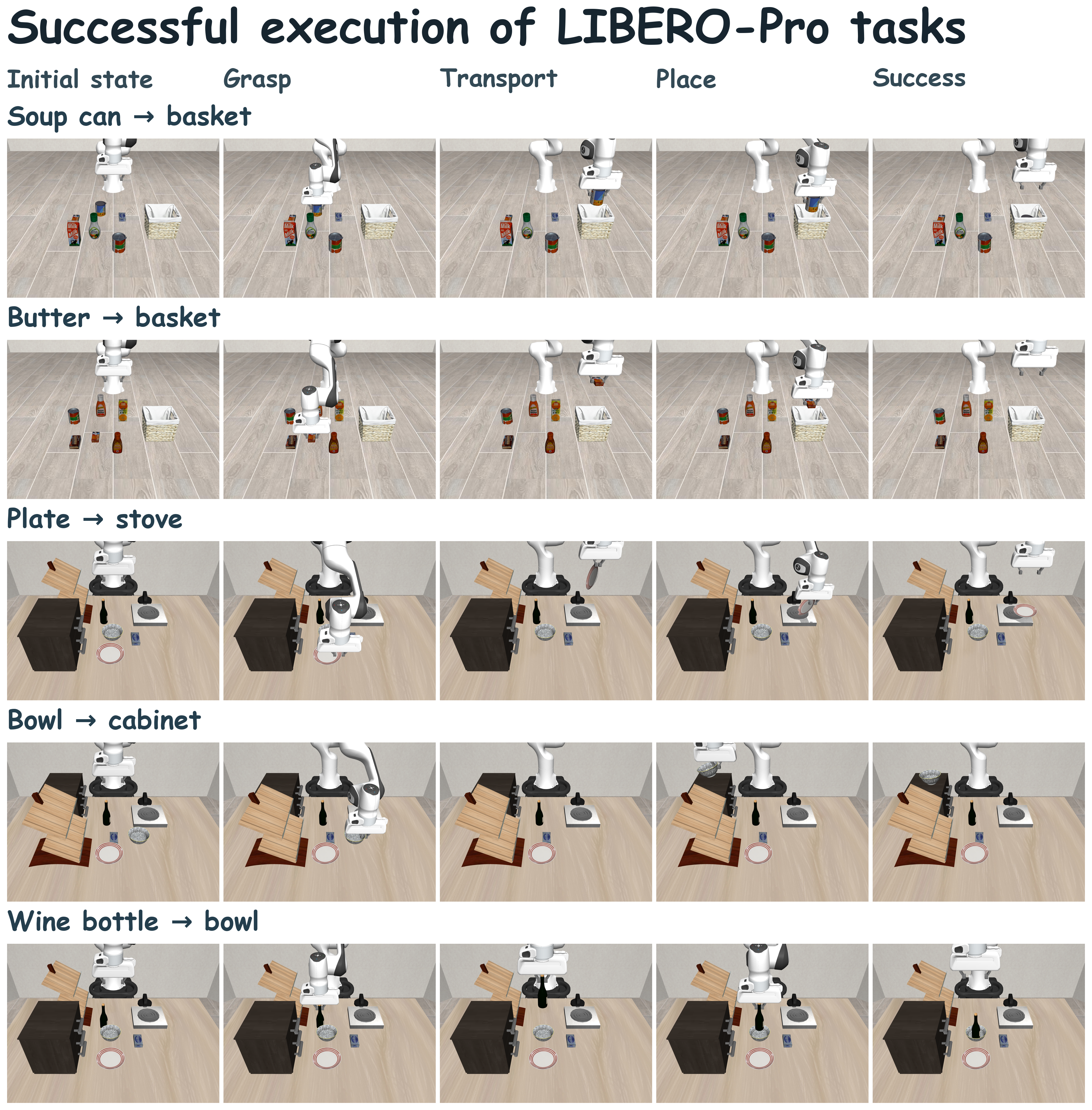}\hfill
  \includegraphics[height=9cm]{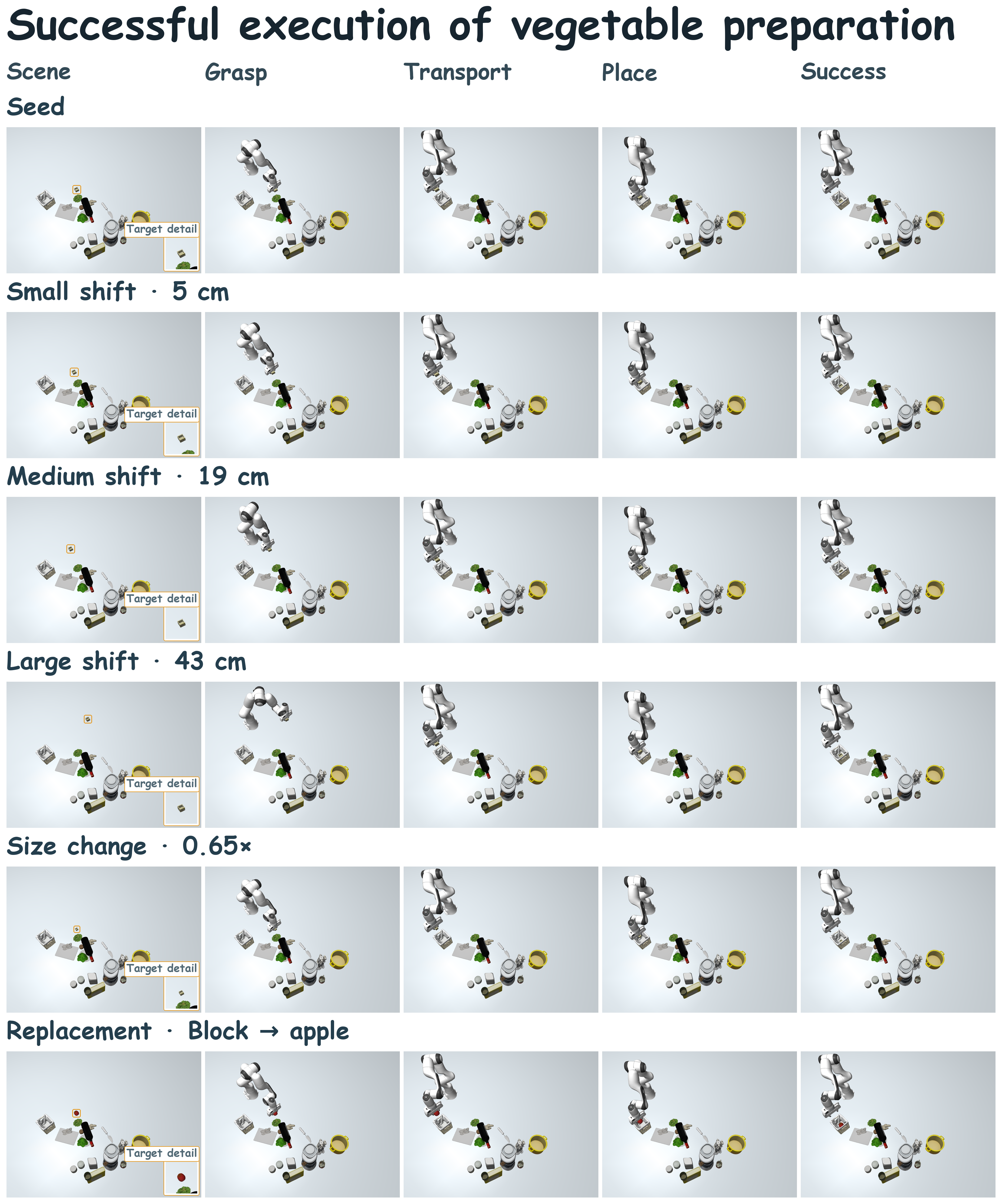}
  \vspace{-0.2cm}
  \caption{\textbf{Successful rollouts.} Left: Pre-defined LIBERO-Pro tasks. Right: Evolved scenes, one row per variant. Columns are the initial state, grasp, transport, place, and success.}
  \label{fig:qual:more}
\end{figure*}

\textbf{Effect of \mempot{}.}
Text2potential achieves
27.6\% OOD success at a fixed training budget, compared with 22.3\% using sparse task rewards alone (\Cref{tab:t2p}).
It reaches 25\% OOD success after 21k policy updates, compared with 31k for sparse rewards, requiring approximately 32\% fewer updates.
The ablations show that both failure diagnosis and potential refinement matter. Generating the potential from the task description instead of
the failure diagnosis lowers success to 24.6\%, while disabling refinement lowers it to 26.1\%.
Keeping the policy frozen and using the value critic only to select action chunks achieves 21.4\%, showing the benefit of using the learned values to update the policy.\looseness-1

\textbf{Qualitative results.}
\Cref{fig:qual:more} shows successful rollouts on LIBERO-Pro tasks and evolved vegetable-preparation scenes. Despite changes in object position, size, and identity, the policy grasps, transports, and places the target, following the subgoal sequence encoded by the staged potential.
Appendix \Cref{fig:qual,fig:qual:libero:more} show how these scenes were modified and provide additional rollouts.\looseness-1

\textbf{Transfer across tasks and policies.}
\Cref{tab:crosstask,tab:crosspolicy} in
Appendix~\ref{app:more} show that transferred generation programs yield greater learning progress
than programs learned from scratch with the same LLM budget, while reusing memory collected with $\pi_{0.5}$ raises OpenVLA-OFT's
OOD success from 3.8\% to 21.1\% without further environment generation. Appendix~\ref{app:more} provides additional results on memory
design and size, training convergence, environment quality,
diagnosis accuracy, computational cost, and generation scheduling.\looseness-1

\section{Conclusion}
\label{sec:conclusion}
We introduce \modelnamenc{}, coupling environment generation and policy learning through shared heterogeneous memory.
Failure diagnoses become reusable programs for generating environments and training the policy.
Text2potential provides dense supervision, while learning progress guides generation and refinement.
Experiments show improved out-of-distribution success, program transfer across task families, and memory reuse across policy backbones.\looseness=-1

\begingroup
\small
\phantomsection
\addcontentsline{toc}{section}{References}
\bibliographystyle{plainnat}
\bibliography{main}

@String(CVPR  = {CVPR})

@article{deitke2023objaverse,
  title={Objaverse-xl: A universe of 10m+ 3d objects},
  author={Deitke, Matt and Liu, Ruoshi and Wallingford, Matthew and Ngo, Huong and Michel, Oscar and Kusupati, Aditya and Fan, Alan and Laforte, Christian and Voleti, Vikram and Gadre, Samir Yitzhak and others},
  journal={Advances in Neural Information Processing Systems},
  volume={36},
  pages={35799--35813},
  year={2023}
}

@article{yu2026dreampartgen,
  title={Dreampartgen: Semantically grounded part-level 3d generation via collaborative latent denoising},
  author={Yu, Tianjiao and Li, Xinzhuo and Wahed, Muntasir and Xiong, Jerry and Shen, Yifan and Shen, Ying and Lourentzou, Ismini},
  journal={arXiv preprint arXiv:2603.19216},
  year={2026}
}

@article{lu2025dreamart,
  title={Dreamart: Generating interactable articulated objects from a single image},
  author={Lu, Ruijie and Liu, Yu and Tang, Jiaxiang and Ni, Junfeng and Wang, Yuxiang and Wan, Diwen and Zeng, Gang and Chen, Yixin and Huang, Siyuan},
  journal={arXiv preprint arXiv:2507.05763},
  year={2025}
}

@inproceedings{yu2026part,
  title={Part\textsuperscript{2}GS: Part-aware Modeling of Articulated Objects using 3D Gaussian Splatting},
  author={Yu, Tianjiao and Shah, Vedant and Wahed, Muntasir and Shen, Ying and Nguyen, Kiet A and Lourentzou, Ismini},
  booktitle={Proceedings of the IEEE/CVF Conference on Computer Vision and Pattern Recognition},
  pages={18913--18923},
  year={2026}
}

@inproceedings{xiang2025structured,
  title={Structured 3d latents for scalable and versatile 3d generation},
  author={Xiang, Jianfeng and Lv, Zelong and Xu, Sicheng and Deng, Yu and Wang, Ruicheng and Zhang, Bowen and Chen, Dong and Tong, Xin and Yang, Jiaolong},
  booktitle={2025 IEEE/CVF Conference on Computer Vision and Pattern Recognition (CVPR)},
  pages={21469--21480},
  year={2025},
  organization={IEEE}
}

@article{zhao2025hunyuan3d,
  title={Hunyuan3d 2.0: Scaling diffusion models for high resolution textured 3d assets generation},
  author={Zhao, Zibo and Lai, Zeqiang and Lin, Qingxiang and Zhao, Yunfei and Liu, Haolin and Yang, Shuhui and Feng, Yifei and Yang, Mingxin and Zhang, Sheng and Yang, Xianghui and others},
  journal={arXiv preprint arXiv:2501.12202},
  year={2025}
}

@article{ranawaka2026simfoundry,
  title   = {SimFoundry: Modular and Automated Scene Generation for Policy Learning and Evaluation},
  author={Ranawaka, Nadun and Wong, Josiah and Pai, Wei-Lin and Chu, Wei-Teng and Dai, Tianyuan and Moghani, Masoud and Yin, Hang and Jiang, Yunfan and Durbano, Wesley and Huynh, Brandon and others},
  journal={arXiv preprint arXiv:2606.28276},
  year={2026}
}

@article{xia2026sage,
  title={Sage: Scalable agentic 3d scene generation for embodied ai},
  author={Xia, Hongchi and Li, Xuan and Li, Zhaoshuo and Ma, Qianli and Xu, Jiashu and Liu, Ming-Yu and Cui, Yin and Lin, Tsung-Yi and Ma, Wei-Chiu and Wang, Shenlong and others},
  journal={arXiv preprint arXiv:2602.10116},
  year={2026}
}

@inproceedings{pfaff2026scenesmith,
  title={Scenesmith: Agentic generation of simulation-ready indoor scenes},
  author={Pfaff, Nicholas and Cohn, Thomas and Zakharov, Sergey and Cory, Rick and Tedrake, Russ},
  booktitle={International Conference on Machine Learning},
  year={2026}
}

@article{wang2023voyager,
  title={Voyager: An open-ended embodied agent with large language models},
  author={Wang, Guanzhi and Xie, Yuqi and Jiang, Yunfan and Mandlekar, Ajay and Xiao, Chaowei and Zhu, Yuke and Fan, Linxi and Anandkumar, Anima},
  journal={arXiv preprint arXiv:2305.16291},
  year={2023}
}

@article{wang2026tableverse,
  title={TableVerse: A Large-scale Tabletop Dataset with Real-world Grounded Layouts for Generalizable Manipulation},
  author={Wang, Boyuan and Zhang, Yue and Xue, Xutao and Song, Xueyu and Sun, Yu},
  journal={arXiv preprint arXiv:2607.21017},
  year={2026}
}

@article{zhou2026articraft,
  title   = {Articraft: An Agentic System for Scalable Articulated 3D Asset Generation},
  author  = {Zhou, Matt and Li, Ruining and Lyu, Xiaoyang and Song, Zhaomou and Huang, Zhening and Zheng, Chuanxia and Rupprecht, Christian and Vedaldi, Andrea and Wu, Shangzhe},
  journal = {arXiv preprint arXiv:2605.15187},
  year    = {2026}
}

@inproceedings{yang2024holodeck,
  title={Holodeck: Language guided generation of 3d embodied ai environments},
  author={Yang, Yue and Sun, Fan-Yun and Weihs, Luca and VanderBilt, Eli and Herrasti, Alvaro and Han, Winson and Wu, Jiajun and Haber, Nick and Krishna, Ranjay and Liu, Lingjie and others},
  booktitle={2024 IEEE/CVF Conference on Computer Vision and Pattern Recognition (CVPR)},
  pages={16277--16287},
  year={2024},
  organization={IEEE}
}

@article{cao2026physxomni,
  title={PhysX-Omni: Unified Simulation-Ready Physical 3D Generation for Rigid, Deformable, and Articulated Objects},
  author={Cao, Ziang and Liu, Yinghao and Li, Haitian and Yao, Runmao and Hong, Fangzhou and Chen, Zhaoxi and Pan, Liang and Liu, Ziwei},
  journal={arXiv preprint arXiv:2605.21572},
  year={2026}
}

@article{yuan2026pace,
  title={PACE: Parameter Change for Unsupervised Environment Design},
  author={Yuan, Fang and Yin, Quanjun and Shen, Siqi and Xie, Yuxiang and Yang, Junqiang and Qin, Long and Zeng, Junjie and Li, Qinglun},
  journal={arXiv preprint arXiv:2605.01358},
  year={2026}
}

@article{xiao2026enpire,
  title={ENPIRE: Agentic Robot Policy Self-Improvement in the Real World},
  author={Xiao, Wenli and Xie, Jia and Zhang, Tonghe and Lin, Haotian and Fu, Letian and Xue, Haoru and Lu, Jalen and Yang, Yi and Dai, Cunxi and Wang, Zi and others},
  journal={arXiv preprint arXiv:2606.19980},
  year={2026}
}

@article{wang2023robogen,
  title={Robogen: Towards unleashing infinite data for automated robot learning via generative simulation},
  author={Wang, Yufei and Xian, Zhou and Chen, Feng and Wang, Tsun-Hsuan and Wang, Yian and Fragkiadaki, Katerina and Erickson, Zackory and Held, David and Gan, Chuang},
  journal={arXiv preprint arXiv:2311.01455},
  year={2023}
}

@inproceedings{xie2024text2reward,
  title={Text2reward: Reward shaping with language models for reinforcement learning},
  author={Xie, Tianbao and Zhao, Siheng and Wu, Chen and Liu, Yitao and Luo, Qian and Zhong, Victor and Yang, Yanchao and Yu, Tao},
  booktitle={International Conference on Learning Representations},
  volume={2024},
  pages={35663--35699},
  year={2024}
}

@article{yu2023language,
  title={Language to rewards for robotic skill synthesis},
  author={Yu, Wenhao and Gileadi, Nimrod and Fu, Chuyuan and Kirmani, Sean and Lee, Kuang-Huei and Arenas, Montse Gonzalez and Chiang, Hao-Tien Lewis and Erez, Tom and Hasenclever, Leonard and Humplik, Jan and others},
  journal={arXiv preprint arXiv:2306.08647},
  year={2023}
}

@inproceedings{ng1999policy,
  title={Policy invariance under reward transformations: Theory and application to reward shaping},
  author={Ng, Andrew Y and Harada, Daishi and Russell, Stuart},
  booktitle={Icml},
  volume={99},
  pages={278--287},
  year={1999} 
}

@article{sumers2023cognitive,
  title={Cognitive architectures for language agents},
  author={Sumers, Theodore R and Yao, Shunyu and Narasimhan, Karthik and Griffiths, Thomas L},
  journal={arXiv preprint arXiv:2309.02427},
  year={2023}
}

@article{wang2025embodiedgen,
  title={Embodiedgen: Towards a generative 3d world engine for embodied intelligence},
  author={Wang, Xinjie and Liu, Liu and Cao, Yu and Wu, Ruiqi and Qin, Wenkang and Wang, Dehui and Sui, Wei and Su, Zhizhong},
  journal={arXiv preprint arXiv:2506.10600},
  year={2025}
}

@inproceedings{le2025articulate,
  title={Articulate-anything: Automatic modeling of articulated objects via a vision-language foundation model},
  author={Le, Long and Xie, Jason and Liang, William and Wang, Hung-Ju and Yang, Yue and Ma, Yecheng Jason and Vedder, Kyle and Krishna, Arjun and Jayaraman, Dinesh and Eaton, Eric},
  booktitle={International Conference on Learning Representations},
  volume={2025},
  pages={17578--17602},
  year={2025}
}

@inproceedings{cao2026physx,
  title={Physx-anything: Simulation-ready physical 3d assets from single image},
  author={Cao, Ziang and Hong, Fangzhou and Chen, Zhaoxi and Pan, Liang and Liu, Ziwei},
  booktitle={Proceedings of the IEEE/CVF Conference on Computer Vision and Pattern Recognition},
  pages={5839--5848},
  year={2026}
}

@article{chen2025robotwin,
  title={Robotwin 2.0: A scalable data generator and benchmark with strong domain randomization for robust bimanual robotic manipulation},
  author={Chen, Tianxing and Chen, Zanxin and Chen, Baijun and Cai, Zijian and Liu, Yibin and Li, Zixuan and Liang, Qiwei and Lin, Xianliang and Ge, Yiheng and Gu, Zhenyu and others},
  journal={arXiv preprint arXiv:2506.18088},
  year={2025}
}

@article{yang2026sceneweaver,
  title={Sceneweaver: All-in-one 3d scene synthesis with an extensible and self-reflective agent},
  author={Yang, Yandan and Jia, Baoxiong and Zhang, Shujie and Huang, Siyuan},
  journal={Advances in neural information processing systems},
  volume={38},
  pages={140319--140351},
  year={2025}
}

@article{wang2026embodiedgen,
  title={EmbodiedGen V2: An Agentic, Simulation-Ready 3D World Engine for Embodied AI},
  author={Wang, Xinjie and Liu, Liu and Ding, Taojun and Choi, Andrew and Huang, Chaodong and Zhao, Mengao and Li, Ziang and Jiang, Jackson and Yu, Chunlei and Liu, Shengxiang and others},
  journal={arXiv preprint arXiv:2607.07459},
  year={2026}
}

@article{dennis2020emergent,
  title={Emergent complexity and zero-shot transfer via unsupervised environment design},
  author={Dennis, Michael and Jaques, Natasha and Vinitsky, Eugene and Bayen, Alexandre and Russell, Stuart and Critch, Andrew and Levine, Sergey},
  journal={Advances in neural information processing systems},
  volume={33},
  pages={13049--13061},
  year={2020}
}

@inproceedings{jiang2021prioritized,
  title={Prioritized level replay},
  author={Jiang, Minqi and Grefenstette, Edward and Rockt{\"a}schel, Tim},
  booktitle={International Conference on Machine Learning},
  pages={4940--4950},
  year={2021},
  organization={PMLR}
}

@inproceedings{liang2024environment,
  title={Environment curriculum generation via large language models},
  author={Liang, William and Wang, Sam and Wang, Hung-Ju and Bastani, Osbert and Jayaraman, Dinesh and Ma, Yecheng Jason},
  booktitle={8th annual conference on robot learning},
  year={2024}
}

@article{li2026reinforcement,
  title={Reinforcement learning with action chunking},
  author={Li, Qiyang and Zhou, Zhiyuan Paul and Levine, Sergey},
  journal={Advances in Neural Information Processing Systems},
  year={2025}
}

@article{giridhar2026beyond,
  title={Beyond Imitation: Self-Improving Robot Policies via Off-Policy Q-Planning},
  author={Giridhar, Varun and Khandelwal, Anant and Collins, Jeremy A and Georgiev, Ignat and Garg, Animesh},
  journal={arXiv preprint arXiv:2608.21204},
  year={2026}
}

@article{kim2024openvla,
  title={Openvla: An open-source vision-language-action model},
  author={Kim, Moo Jin and Pertsch, Karl and Karamcheti, Siddharth and Xiao, Ted and Balakrishna, Ashwin and Nair, Suraj and Rafailov, Rafael and Foster, Ethan and Sanketi, Pannag and others},
  journal={Conference on Robot Learning (CoRL)},
  year={2024}
}

@inproceedings{yao2025think,
  title={Think small, act big: Primitive prompt learning for lifelong robot manipulation},
  author={Yao, Yuanqi and Liu, Siao and Song, Haoming and Qu, Delin and Chen, Qizhi and Ding, Yan and Zhao, Bin and Wang, Zhigang and Li, Xuelong and Wang, Dong},
  booktitle={2025 IEEE/CVF Conference on Computer Vision and Pattern Recognition (CVPR)},
  pages={22573--22583},
  year={2025},
  organization={IEEE}
}

@article{lu2026aspire,
  title={ASPIRE: Agentic/Skills Discovery for Robotics},
  author={Lu, Runyu and Wu, Yubo and Kou, Ethan and Fu, Letian and Xiao, Wenli and Mandlekar, Ajay and Xu, Yinzhen and Shi, Guanya and Goldberg, Ken and Chen, Ang and others},
  journal={arXiv preprint arXiv:2607.00272},
  year={2026}
}

@article{kang2026simworld,
  title={SimWorld Studio: Automatic Environment Generation with Evolving Coding Agent for Embodied Agent Learning},
  author={Kang, Haoqiang and Ye, Xiaokang and Liu, Yuhan and Mantri, Siddhant Hitesh and Mao, Lingjun and Fleming, James and Regmi, Drishti and Qin, Lianhui},
  journal={arXiv preprint arXiv:2605.09423},
  year={2026}
}

@article{wang2019paired,
  title={Paired open-ended trailblazer (poet): Endlessly generating increasingly complex and diverse learning environments and their solutions},
  author={Wang, Rui and Lehman, Joel and Clune, Jeff and Stanley, Kenneth O},
  journal={arXiv preprint arXiv:1901.01753},
  year={2019}
}

@inproceedings{parker2022evolving,
  title={Evolving curricula with regret-based environment design},
  author={Parker-Holder, Jack and Jiang, Minqi and Dennis, Michael and Samvelyan, Mikayel and Foerster, Jakob and Grefenstette, Edward and Rockt{\"a}schel, Tim},
  booktitle={International Conference on Machine Learning},
  pages={17473--17498},
  year={2022},
  organization={PMLR}
}

@article{portelas2020automatic,
  title={Automatic curriculum learning for deep rl: A short survey},
  author={Portelas, R{\'e}my and Colas, C{\'e}dric and Weng, Lilian and Hofmann, Katja and Oudeyer, Pierre-Yves},
  journal={arXiv preprint arXiv:2003.04664},
  year={2020}
}

@inproceedings{ma2024eureka,
  title={Eureka: Human-level reward design via coding large language models},
  author={Ma, Yecheng Jason and Liang, William and Wang, Guanzhi and Huang, De-An and Bastani, Osbert and Jayaraman, Dinesh and Zhu, Yuke and Fan, Jim and others},
  booktitle={International conference on learning Representations},
  volume={2024},
  pages={26516--26560},
  year={2024}
}

@article{peng2019advantage,
  title={Advantage-weighted regression: Simple and scalable off-policy reinforcement learning},
  author={Peng, Xue Bin and Kumar, Aviral and Zhang, Grace and Levine, Sergey},
  journal={arXiv preprint arXiv:1910.00177},
  year={2019}
}

@inproceedings{li2026simplevla,
  title={Simplevla-rl: Scaling vla training via reinforcement learning},
  author={Li, Haozhan and Zuo, Yuxin and Yu, Jiale and Zhang, Yuhao and Yang, Zhaohui and Zhang, Kaiyan and Zhu, Xuekai and Zhang, Yuchen and Chen, Tianxing and Cui, Ganqu and others}, 
  booktitle={International Conference on Learning Representations},
  volume={2026},
  pages={125846--125869},
  year={2026}
}

@article{tan2025interactive,
  title={Interactive post-training for vision-language-action models},
  author={Tan, Shuhan and Dou, Kairan and Zhao, Yue and Kr{\"a}henb{\"u}hl, Philipp},
  journal={arXiv preprint arXiv:2505.17016},
  year={2025}
}

@article{kim2025fine,
  title={Fine-tuning vision-language-action models: Optimizing speed and success},
  author={Kim, Moo Jin and Finn, Chelsea and Liang, Percy},
  journal={arXiv preprint arXiv:2502.19645},
  year={2025}
}

@article{black2024pi_0,
  title={$\pi_0$: A Vision-Language-Action Flow Model for General Robot Control},
  author={Black, Kevin and Brown, Noah and Driess, Danny and Esmail, Adnan and Equi, Michael and Finn, Chelsea and Fusai, Niccolo and Groom, Lachy and Hausman, Karol and Ichter, Brian and others},
  journal={arXiv preprint arXiv:2410.24164},
  year={2024}
}

@article{zhou2025libero,
  title={Libero-pro: Towards robust and fair evaluation of vision-language-action models beyond memorization},
  author={Zhou, Xueyang and Xu, Yangming and Tie, Guiyao and Chen, Yongchao and Zhang, Guowen and Chu, Duanfeng and Zhou, Pan and Sun, Lichao},
  journal={arXiv preprint arXiv:2510.03827},
  year={2025}
}

@article{intelligence2025pi_,
  title={$\pi_{0.5}$: a Vision-Language-Action Model with Open-World Generalization},
  author={Intelligence, Physical and Black, Kevin and Brown, Noah and Darpinian, James and Dhabalia, Karan and Driess, Danny and Esmail, Adnan and Equi, Michael and Finn, Chelsea and Fusai, Niccolo and others},
  journal={arXiv preprint arXiv:2504.16054},
  year={2025}
}

@article{oquab2023dinov2,
  title={Dinov2: Learning robust visual features without supervision},
  author={Oquab, Maxime and Darcet, Timoth{\'e}e and Moutakanni, Th{\'e}o and Vo, Huy and Szafraniec, Marc and Khalidov, Vasil and Fernandez, Pierre and Haziza, Daniel and Massa, Francisco and El-Nouby, Alaaeldin and others},
  journal={arXiv preprint arXiv:2304.07193},
  year={2023}
}

@article{raffel2020exploring,
  title={Exploring the limits of transfer learning with a unified text-to-text transformer},
  author={Raffel, Colin and Shazeer, Noam and Roberts, Adam and Lee, Katherine and Narang, Sharan and Matena, Michael and Zhou, Yanqi and Li, Wei and Liu, Peter J},
  journal={Journal of machine learning research},
  volume={21},
  number={140},
  pages={1--67},
  year={2020}
}

@article{sundaralingam2023curobo,
  title={curobo: Parallelized collision-free minimum-jerk robot motion generation},
  author={Sundaralingam, Balakumar and Hari, Siva Kumar Sastry and Fishman, Adam and Garrett, Caelan and Van Wyk, Karl and Blukis, Valts and Millane, Alexander and Oleynikova, Helen and Handa, Ankur and Ramos, Fabio and others},
  journal={arXiv preprint arXiv:2310.17274},
  year={2023}
}

@misc{openpi2025,
  title        = {openpi: Open-Source Models and Code from {Physical Intelligence}},
  author       = {{Physical Intelligence}},
  year         = {2025},
  howpublished = {\url{https://github.com/Physical-Intelligence/openpi}},
  note         = {Accessed: 2026-09-26}
}

@misc{openai2026gpt55,
  author       = {{OpenAI}},
  title        = {{GPT-5.5 System Card}},
  year         = {2026},
  howpublished = {\url{https://openai.com/index/gpt-5-5-system-card/}}
}
\endgroup

\clearpage
\appendix
\newpage
\section{Transferring Generation Programs Across Task Families}
\label{app:lifting}

A generation program can be reused in a new task family when its object
requirements and relational constraints are satisfied in the new scene.
We call this transfer \emph{lifting}. Each program refers to objects by
symbolic role, such as \texttt{target}, \texttt{obstacle}, or
\texttt{support}, rather than by a specific object instance. Each role
specifies a required object category, such as \emph{container}, or an
affordance, such as \emph{graspable}. The program also specifies the
relations required between these roles, such as an obstacle occupying
the approach path to the target.\looseness=-1

To apply a program to a new scene, $\Rgrind$ matches these requirements
against the scene program's symbol table, which records object
identifiers, categories, affordances, poses, and relations. This
information is available directly because the data engine constructs
the scene program. A valid \emph{binding} assigns each role to a
distinct scene object and satisfies all role requirements and
relational constraints. If no valid binding exists, the program is
not transferred to that scene. Matching is exhaustive: our programs
contain at most five roles and our scenes contain a few dozen
objects, with matching taking milliseconds in this setting.

When several bindings are valid, they are ranked first by how closely
their object categories match those in the difficulty mode's
previously validated variants. Ties are broken using the margins
by which the relational constraints are satisfied. We retain at most
three bindings and execute the generation program separately for
each, producing a scene variant and its own matched control batch.\looseness-1

Transferred programs undergo the same validation as programs learned
within the new task family. Their generated variants must pass the
scene-validity checks, and policy rollouts evaluate both the targeted
variants and their controls. A transferred program enters $\skpool$
for the new family only when the control comparison supports the
attributed difficulty and the measured success rates of its variants
fall within the success band. Thus, matching establishes whether
a program is applicable to the new scene; evaluation establishes
whether it reproduces a useful training difficulty.

\section{Implementation Details}
\label{app:impl}

\textbf{Environments and evaluation.}
\label{sec:exp:setup}
We construct MuJoCo environments for four task families:
\emph{pick-and-place}, \emph{articulated open/close},
\emph{put-inside}, and \emph{stack/insert}. Each family includes
an in-distribution evaluation suite and four held-out suites of
100 environments each, testing generalization to unseen object
instances, layouts, tasks, and categories, respectively.
Held-out environments never enter $\mem$.
In a second setting, we initialize the loop with the 30 LIBERO
source tasks and evaluate on LIBERO-Pro
\citep{zhou2025libero} (\Cref{tab:liberopro}).
All results are reported as mean $\pm$ standard error over
three random seeds.\looseness-1

\textbf{Assets, observations, and actions.}
The asset collection comprises 1{,}200 rigid objects from
Objaverse across 60 categories, 300 articulated objects from
ArtiCraft-10K, and 500 seed layouts from TableVerse
\citep{deitke2023objaverse, zhou2026articraft, wang2026tableverse}.
The robot is a 7-DoF Franka Panda with front and wrist cameras,
each providing $224\times224$ RGB observations.
Control runs at 10\,Hz using seven-dimensional actions:
six dimensions for the change in end-effector pose and one
for the gripper command. Each action chunk contains $h=8$
control steps.

\textbf{Policy backbones.}
\label{sec:exp:policies}
Our primary policy is $\pi_{0.5}$, a 3B-parameter model with
a flow-matching action head \citep{intelligence2025pi_}.
We initialize it from the released openpi base checkpoint,
or from its LIBERO-fine-tuned checkpoint for the LIBERO
experiments \citep{openpi2025}.
We also evaluate the 7B-parameter OpenVLA-OFT
\citep{kim2025fine} in \cref{tab:crosspolicy}.
Both policies are fine-tuned using LoRA with rank 32 and
a learning rate of $10^{-5}$.
At each rollout decision, the policy samples $j\!=\!8$
candidate action chunks. These are combined using softmax
weights computed from $\Qf$, with temperature $0.5$.\looseness-1

\textbf{Scene generation, diagnosis, and potential compilation.}
The data engine, rollout-diagnosis model, and potential
compiler all use GPT-5.5~\citep{openai2026gpt55}. Planner-certified demonstrations
are obtained using cuRobo~\citep{sundaralingam2023curobo}.
The diagnosis model receives rollout video, contact
information, joint states, and the task success verifier's
output at each timestep. It assigns each diagnosed failure
to one of four factors
(\texttt{TASK}, \texttt{SPATIAL}, \texttt{ATTRIBUTE}, or
\texttt{EXECUTION}) and one of five trajectory phases
(approach, grasp, transport, place or insert, and release).
For potential compilation, \mempot{} retrieves three
exemplar programs from $\dirpool_j$ and uses a library of
12 helper functions. Generated programs undergo at most
two rounds of repair for syntax or runtime errors. \looseness-1

\textbf{Value learning and policy optimization.}
The value critic $\Qf$ uses frozen DINOv2-B~\citep{oquab2023dinov2} and T5-base~\citep{raffel2020exploring}
encoders followed by a four-layer transformer decoder.
It is trained by categorical value regression with
101 bins spanning $[-\beta m_{\max},1]$, where $\beta=0.5$
is the shaping coefficient and $m_{\max}=4$ is the
maximum number of subgoals.
We use a discount factor of $\gamma\!=\!0.99$ and an exponential
moving average target network with update coefficient
$\tau\!=\!0.005$.
Each round includes $K_Q\!=\!20{,}000$ critic-update steps
with batch size 256 and $K_\pi\!=\!5{,}000$ policy-update steps
with batch size 64. The advantage-weighted policy update
uses $\lambda\!=\!1$.\looseness-1

\begin{table*}[t]
  \centering\small
  \caption{\textbf{Effect of memory design on policy learning.}
OOD success (\%), mean $\pm$ standard error over three seeds
at the same fixed generation budget, and \# rounds required to
reach 25\% OOD success.
$\Omega$ and $\Phi$ denote scene generation and policy
training programs, respectively.
The full method retains both program types.
   \protect\captionbox{CoDABest}{\textbf{Best}} and \protect\captionbox{CoDASecond}{second} highlighted.}
  \label{tab:fork}
  \vspace{-0.2cm}
  \begin{tabularx}{\linewidth}{@{}>{\raggedright\arraybackslash}Xcc@{}}
    \toprule
    \textbf{Memory configuration} & \textbf{OOD success} & \textbf{Rounds to target} \\
    \midrule
    No memory                                            & 19.2\pmse{0.7} & 14 \\
    $\skpool$ only (engine side)                         & 22.8\pmse{0.7} & 9 \\
    $\dirpool$ only (policy side)                        & 22.1\pmse{0.8} & 11 \\
    Single memory: raw trajectory retrieval               & 20.1\pmse{0.9} & 13 \\
    Single memory: distilled insights                     & 21.6\pmse{0.7} & 12 \\
    Single memory: skill library \citep{wang2023voyager}  & 21.0\pmse{0.8} & 12 \\
    $\dirpool$ with free-form reward (not potential)     & 24.3\pmse{0.9} & 8 \\
    $\skpool$ without lifting (task-bound programs)      & \second{25.4\pmse{0.6}} & \second{7} \\
    \midrule
    \rowcolor{gray!8}
    \textbf{Forked (ours)}                               & \best{27.6\pmse{0.5}} & \best{5} \\
    \bottomrule
  \end{tabularx}

\end{table*}

\begin{table*}[t]
  \centering\small
  \caption{\textbf{Transfer of generation programs across task families.}
Rows indicate the family in which programs were learned, and
columns indicate the family in which they are evaluated.
Each entry reports two percentages:
variants within the success band / learning progress relative
to programs learned within the target family.
Diagonal entries provide the within-family reference,
with learning progress normalized to 100\%.
The final row reports programs learned from scratch in the
target family using the same LLM token budget as transfer.\looseness=-1}

  \label{tab:crosstask}
  \vspace{-0.2cm}
  \begin{tabularx}{\linewidth}{@{}>{\raggedright\arraybackslash}Xcccc@{}}
    \toprule
    \textbf{Learned on $\downarrow$ / lifted to $\rightarrow$} & \textbf{Pick-place} & \textbf{Articulated} & \textbf{Put-inside} & \textbf{Stack/insert} \\
    \midrule
    Pick-place   & 72 / 100 & 38 / 46 & 61 / 79 & 55 / 68 \\
    Articulated  & 41 / 52 & 69 / 100 & 44 / 55 & 37 / 48 \\
    Put-inside   & 63 / 81 & 40 / 50 & 70 / 100 & 52 / 64 \\
    Stack/insert & 57 / 70 & 36 / 45 & 54 / 66 & 66 / 100 \\
    \midrule
    Re-learned from scratch on target & 31 / 38 & 28 / 35 & 30 / 37 & 27 / 33 \\
    \bottomrule
  \end{tabularx}
\end{table*}

\textbf{Generation budgets and rollout protocol.}
Experiments use $n\!=\!5$ rounds, with per-family budgets of 320 accepted environments, 1{,}280 planner-certified demonstrations, and 2.5M LLM tokens. Token counts include scene generation, diagnosis, and
potential compilation. Extended runs use $n\!=\!15$ rounds.
Each round generates $B\!=\!64$ targeted environments and
64 control environments per task family.
The policy is evaluated on each environment using $N\!=\!16$
rollouts from the same initial simulator state.
The diagnosis model processes at most $C\!=\!32$ failed
rollouts per difficulty mode.
The success band is $[0.2,0.8]$.
We reserve a fraction $\varepsilon\!=\!0.15$ of the generation
budget for novel environments and $0.10$ for replay,
and use an upper-confidence-bound exploration coefficient
of $1$ when selecting difficulty modes.

\textbf{Round scheduling and randomization.}
A new round is triggered when fewer than eight difficulty
modes remain within the success band, or when success on
modes within the band changes by less than two percentage
points over the preceding $2{,}000$ policy-update steps.
The three random seeds jointly vary environment generation,
policy rollouts, and model training.

\section{Additional Experiments}
\label{app:more}

\textbf{Complementary generation and training programs.}
\label{sec:exp:hetero}
\Cref{tab:fork} evaluates the contributions of the generation
and training programs retained in memory.
Retaining only scene generation programs raises OOD success
from 19.2\% to 22.8\%, while retaining only policy training
programs raises it to 22.1\%.
Retaining both reaches 27.6\%: the resulting
8.4-percentage-point improvement exceeds the sum of the
individual gains, 3.6 and 2.9 points.
The three baseline variants that reuse the same memory representation
for both scene generation and policy training perform worse
than retaining either type of specialized program alone.
With both generation and training programs, the policy
reaches 25\% OOD success in five rounds, compared with
fourteen rounds without memory.\looseness-1

\begin{table*}[t]
  \centering\small
  \caption{\textbf{Transfer of shared memory across policy backbones.}
OOD success (\%), with rows identifying the backbone used
to collect memory and columns identifying the backbone
trained and evaluated.
Off-diagonal entries use frozen memory from the other
backbone, with no additional environment generation.
Diagonal entries report the full \modelnamenc{} framework
run with the corresponding backbone.
The final row gives each backbone's SFT baseline without
memory.  \protect\captionbox{CoDABest}{\textbf{Best}} and \protect\captionbox{CoDASecond}{second best} results in each column
are highlighted.}
  \label{tab:crosspolicy}
  \vspace{-0.2cm}
  \begin{tabularx}{\linewidth}{@{}>{\raggedright\arraybackslash}Xcc@{}}
    \toprule
    \textbf{Loop run around $\downarrow$ / policy trained $\rightarrow$} & {$\pi_{0.5}$} & {OpenVLA-OFT} \\
    \midrule
    $\pi_{0.5}$        & \best{27.6} & \second{21.1} \\
    OpenVLA-OFT        & \second{25.8} & \best{22.7} \\
    \midrule
    no memory & 5.3 & 3.8 \\
    \bottomrule
  \end{tabularx}

\end{table*}

\begin{table*}[tp]
  \centering\small\setlength{\tabcolsep}{4pt}
  \caption{\textbf{Generation quality and validation of failure diagnoses.}
Verification (\%) measures the percentage of scene proposals passing
physical validity checks.
Coverage is the entropy over task, spatial, and object-attribute
factors $(\tau,\sigma,\alpha)$.
Targeting measures the percentage of generated variants
within the success band. Refuted reports the percentage of failure diagnoses rejected after comparing policy performance on targeted and control scene variants.
 \protect\captionbox{CoDABest}{\textbf{Best}} and \protect\captionbox{CoDASecond}{second best} highlighted.}
  \label{tab:gen}
  \vspace{-0.2cm}
  \begin{tabularx}{\linewidth}{@{}>{\raggedright\arraybackslash}Xccccc@{}}
    \toprule
    & \shortstack{\textbf{Verification}\\(\%) $\uparrow$} & \shortstack{\textbf{Coverage}\\(bits) $\uparrow$} & \shortstack{\textbf{Targeting}\\(\%) $\uparrow$} & \shortstack{\textbf{Refuted}\\(\%)} & \shortstack{\textbf{Human agr.}\\(\%) $\uparrow$} \\
    \midrule
    uniform variants      & 71.3 & \best{3.42} & 24.6 & --- & --- \\
    \modelnamenc{}, no controls    & \second{88.2} & \second{3.21} & \second{58.4} & --- & \second{73.6} \\
    \rowcolor{gray!8}
    \modelnamegradient{} (ours) & \best{88.9} & 3.18 & \best{66.7} & 18.2 & \best{82.4} \\
    \bottomrule
  \end{tabularx}
\end{table*}

\begin{table*}[t]
  \centering\small\setlength{\tabcolsep}{4pt}
  \caption{\textbf{Computational cost and effect of round scheduling.}
LLM token usage, simulation time, and GPU time are reported
per 1{,}000 accepted environments. Token counts include scene generation, failure diagnosis, and potential compilation.
The fixed schedule starts a new generation round every
$K_\pi$ policy-update steps, while the adaptive schedule
uses the remaining difficulty modes and recent learning
progress. \protect\captionbox{CoDABest}{\textbf{Best}} and \protect\captionbox{CoDASecond}{second best} highlighted.\looseness=-1}
  \label{tab:cost}
   \vspace{-0.2cm}
  \begin{tabularx}{\linewidth}{@{}>{\raggedright\arraybackslash}Xcccc@{}}
    \toprule
    & \shortstack{\textbf{LLM tokens}\\(M) $\downarrow$} & \shortstack{\textbf{Sim-hours}\\$\downarrow$} & \shortstack{\textbf{GPU-hours}\\$\downarrow$} & \shortstack{\textbf{OOD success}\\$\uparrow$} \\
    \midrule
    uniform variants         & \best{3.1} & \best{14.2} & \best{38} & 8.5 \\
    scalar-feedback          & \second{4.6} & \second{22.5} & \second{51} & 18.1 \\
    \modelnamegradient{}, fixed schedule       & 8.3 & 27.8 & 74 & \second{26.1} \\
    \rowcolor{gray!8}
    \modelnamegradient{}, adaptive schedule  & 7.4 & 25.1 & 66 & \best{27.6} \\
    \bottomrule
  \end{tabularx}
\vspace{0.1em}

\end{table*}

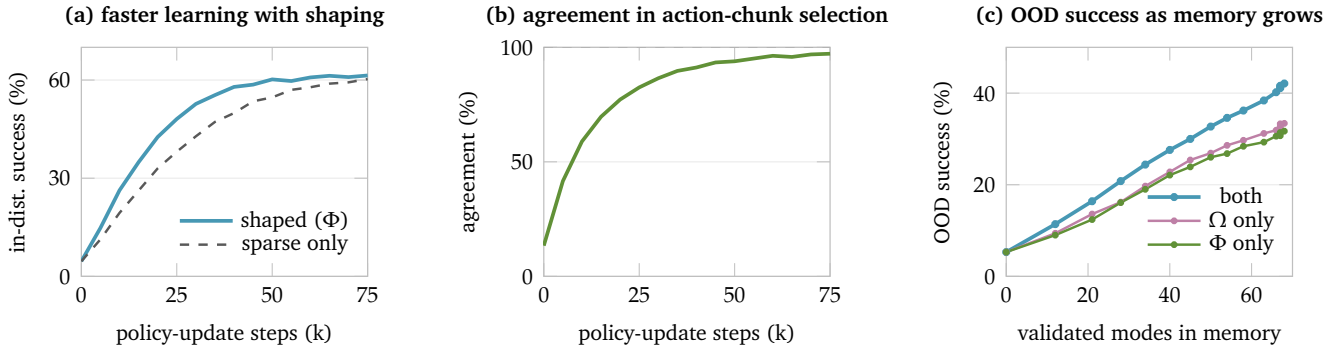
\begin{figure*}[!t]
  \centering
  \resizebox{\linewidth}{!}{\begin{tikzpicture}
\pgfplotsset{every axis/.append style={scale only axis, width=3.25cm, height=2.6cm,
  tick label style={font=\scriptsize}, label style={font=\scriptsize},
  title style={font=\scriptsize\bfseries, yshift=-0.6ex}, axis line style={gray!50}, tick style={gray!50},
  ymajorgrids, grid style={gray!15}, legend style={font=\scriptsize, draw=none, fill=none, inner sep=1pt, row sep=-3pt}}}
\begin{groupplot}[group style={group size=3 by 1, horizontal sep=2cm}]
\nextgroupplot[title={(a) faster learning with shaping}, xlabel={policy-update steps (k)}, ylabel={in-dist.\ success (\%)},
  xmin=0, xmax=75, ymin=0, ymax=70, xtick={0,25,50,75}, ytick={0,30,60}, legend style={at={(0.97,0.05)}, anchor=south east}]
\addplot[WhetBlue!80!black, very thick] coordinates {(0,4.6) (5,14.8) (10,26.3) (15,34.9) (20,42.6) (25,48.1) (30,52.7) (35,55.4) (40,57.9) (45,58.6) (50,60.2) (55,59.7) (60,60.8) (65,61.3) (70,60.9) (75,61.4)};
\addlegendentry{shaped ($\pot$)}
\addplot[gray!70!black, thick, dashed] coordinates {(0,4.6) (5,11.3) (10,19.4) (15,26.2) (20,33.0) (25,38.1) (30,42.8) (35,47.2) (40,49.9) (45,53.5) (50,54.7) (55,57.0) (60,57.8) (65,58.9) (70,59.3) (75,60.3)};
\addlegendentry{sparse only}
\nextgroupplot[title={(b) agreement in action-chunk selection}, xlabel={policy-update steps (k)}, ylabel={agreement (\%)},
  xmin=0, xmax=75, ymin=0, ymax=100, xtick={0,25,50,75}, ytick={0,50,100}]
\addplot[WhetGreen!70!black, very thick] coordinates {(0,13.4) (5,41.6) (10,58.9) (15,69.7) (20,77.2) (25,82.5) (30,86.5) (35,89.7) (40,91.2) (45,93.4) (50,93.9) (55,95.1) (60,96.3) (65,95.8) (70,96.9) (75,97.2)};
\draw[gray!60, dashed] (axis cs:0,100) -- (axis cs:75,100);
\nextgroupplot[title={(c) OOD success as memory grows}, xlabel={validated modes in memory}, ylabel={OOD success (\%)},
  xmin=0, xmax=70, ymin=0, ymax=50, xtick={0,20,40,60}, ytick={0,20,40}, legend style={at={(0.97,0.05)}, anchor=south east}]
\addplot[WhetBlue!80!black, very thick, mark=*, mark size=0.7pt] coordinates {(0,5.3) (12,11.4) (21,16.4) (28,20.8) (34,24.4) (40,27.6) (45,30.0) (50,32.7) (54,34.6) (58,36.2) (63,38.4) (66,40.2) (67,41.1) (67,41.4) (67,41.6) (68,42.1)};
\addlegendentry{both}
\addplot[WhetPink!80!black, thick, mark=*, mark size=0.7pt] coordinates {(0,5.3) (12,9.4) (21,13.6) (28,16.2) (34,19.7) (40,22.8) (45,25.4) (50,26.9) (54,28.6) (58,29.7) (63,31.2) (66,31.9) (67,32.7) (67,32.6) (67,33.3) (68,33.4)};
\addlegendentry{$\skpool$ only}
\addplot[WhetGreen!70!black, thick, mark=*, mark size=0.7pt] coordinates {(0,5.3) (12,9.0) (21,12.4) (28,16.1) (34,19.0) (40,22.1) (45,23.9) (50,26.0) (54,26.8) (58,28.4) (63,29.3) (66,30.6) (67,30.7) (67,31.4) (67,31.2) (68,31.7)};
\addlegendentry{$\dirpool$ only}
\end{groupplot}
\end{tikzpicture}}
  \vspace{-0.6cm}
  \caption{\textbf{Learning dynamics and memory accumulation.}
  (a) In-distribution success on pick-and-place tasks during
  policy training, using the same generated data with
  potential shaping or sparse rewards alone. Shaping
  accelerates learning, with both conditions reaching similar
  final success rates.
  (b) Agreement between the action chunks selected by the
  two value critics, reaching approximately 97\% at the end
  of training.
  (c) OOD success at successive memory snapshots, indexed
  by the number of validated difficulty modes. Using both
  program stores yields higher final success than retaining
  generation or training programs alone.\looseness=-1}
  \label{fig:invariance}
  \label{fig:memsize}
\end{figure*}

\textbf{Transfer across task families.}
\Cref{tab:crosstask} evaluates generation programs transferred
to a different task family by rebinding their object roles,
without relearning the programs. 
The baseline learns programs from scratch in the target
family using the same LLM token budget as transfer. Averaged across all cross-family transfers, reused programs
achieve 60.3\% of the learning progress obtained with
programs learned within the target family, compared with
35.8\% for the from-scratch baseline.
This advantage holds in every cross-family comparison,
demonstrating that generation experience can be reused
to support policy learning in new task families.\looseness=-1

\textbf{Transfer across policy backbones.}
\Cref{tab:crosspolicy} evaluates whether memory collected
with one policy backbone can support training another,
with the memory held fixed and no additional environment
generation.
Transferred memory improves OOD success by an average of
18.9 percentage points over the SFT baseline without memory.
In both transfer directions, performance remains within
two percentage points of running the full framework with
the target backbone.
These results show that the accumulated difficulty modes
and programs can support policy learning beyond the backbone
used to collect them.\looseness-1

\textbf{Learning speed, action agreement, and performance as memory accumulates.}
On the same generated data, potential shaping accelerates
policy learning relative to sparse-only training
(\Cref{fig:invariance}(a,b)).
Both conditions reach similar final in-distribution success
rates, but shaping reaches this level approximately
20{,}000 policy-update steps earlier.
At the end of training, the two value critics select the
same action chunk in 97\% of comparisons.
These results support the practical benefit of the shaping
method: faster learning
with closely aligned final action preferences.
\Cref{fig:invariance}(c) tracks OOD success at successive
memory snapshots during training.
Success increases as more difficulty modes are validated
and recorded.
Retaining both scene generation and policy training programs
achieves the highest final success, approximately ten
percentage points above retaining either program type alone.\looseness-1

\textbf{Generation quality and diagnosis validation.}
\label{sec:exp:gen}
\Cref{tab:gen} shows that 88.9\% of scene proposals pass
the physical validity checks, compared with 71.3\% under
uniform sampling.
Coverage entropy remains close to that of uniform sampling
(3.18 versus 3.42 bits), while targeting precision increases
from 24.6\% to 66.7\%.
The engine therefore improves scene validity and targeting
while maintaining broad coverage.
We reject 18.2\% of the critic's failure diagnoses because
variants that modify the identified scene factor are no harder
than control variants that modify a different factor.
In the evaluation with two annotators and 100 sampled
diagnoses, this validation step improves agreement with
human annotations from 73.6\% to 82.4\%.

\textbf{Computational cost and adaptive scheduling.}
\label{sec:exp:cost}
\modelnamenc{} improves OOD success from 18.1\% to 27.6\%
over scalar feedback, a 52\% relative improvement
(\Cref{tab:cost}).
Its computational requirements are 7.4M LLM tokens and
66 GPU-hours per 1{,}000 accepted environments.
Adaptive scheduling improves efficiency, reducing
both token usage and GPU time by approximately 11\%
relative to fixed round intervals while increasing
OOD success from 26.1\% to 27.6\%.
Thus, adapting when to generate new environments reduces
computational cost while improving policy performance.

\begin{figure*}[t]
  \centering
  \vspace{-0.2cm}
  \includegraphics[width=0.48\linewidth]{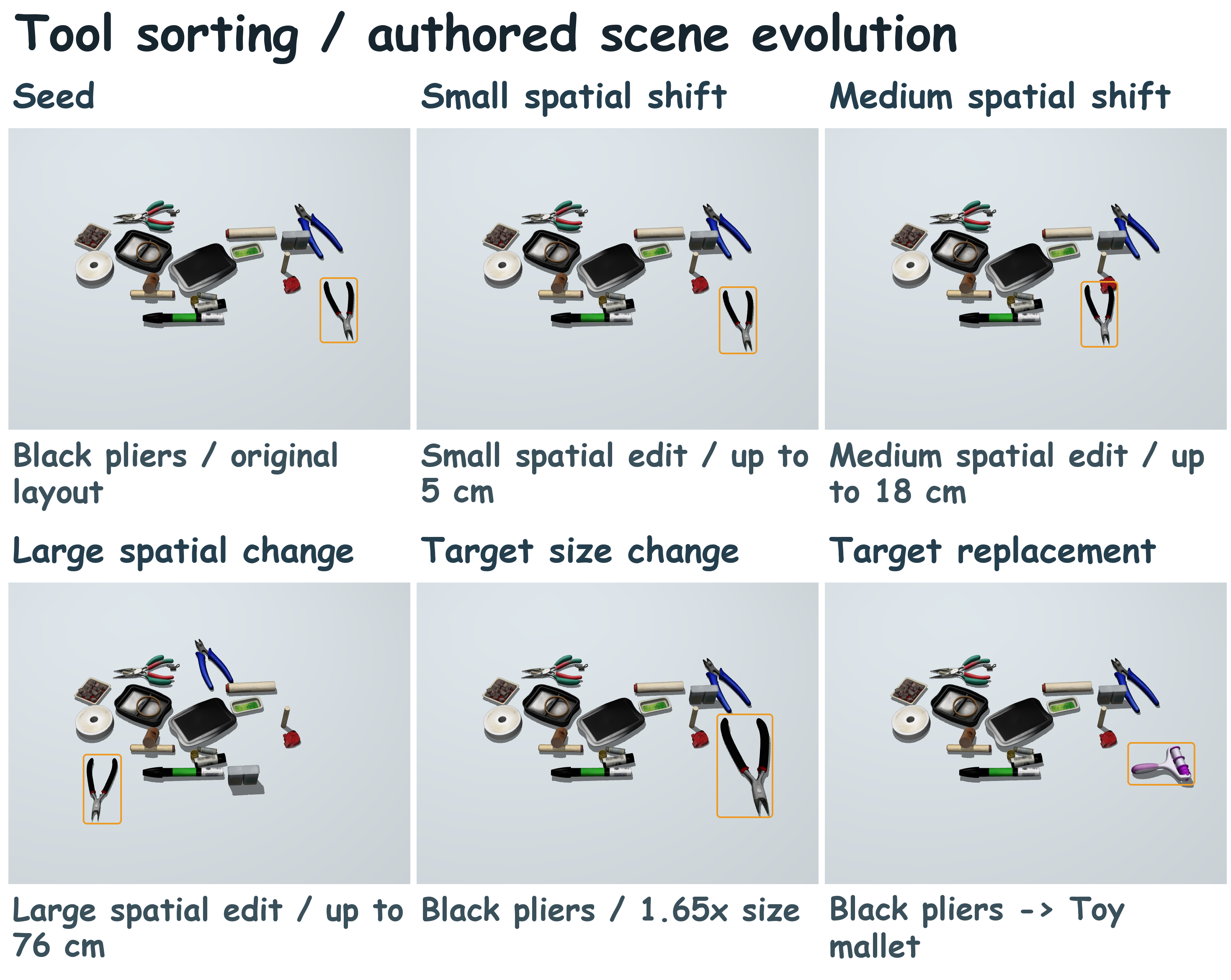}\hfill
  \includegraphics[width=0.48\linewidth]{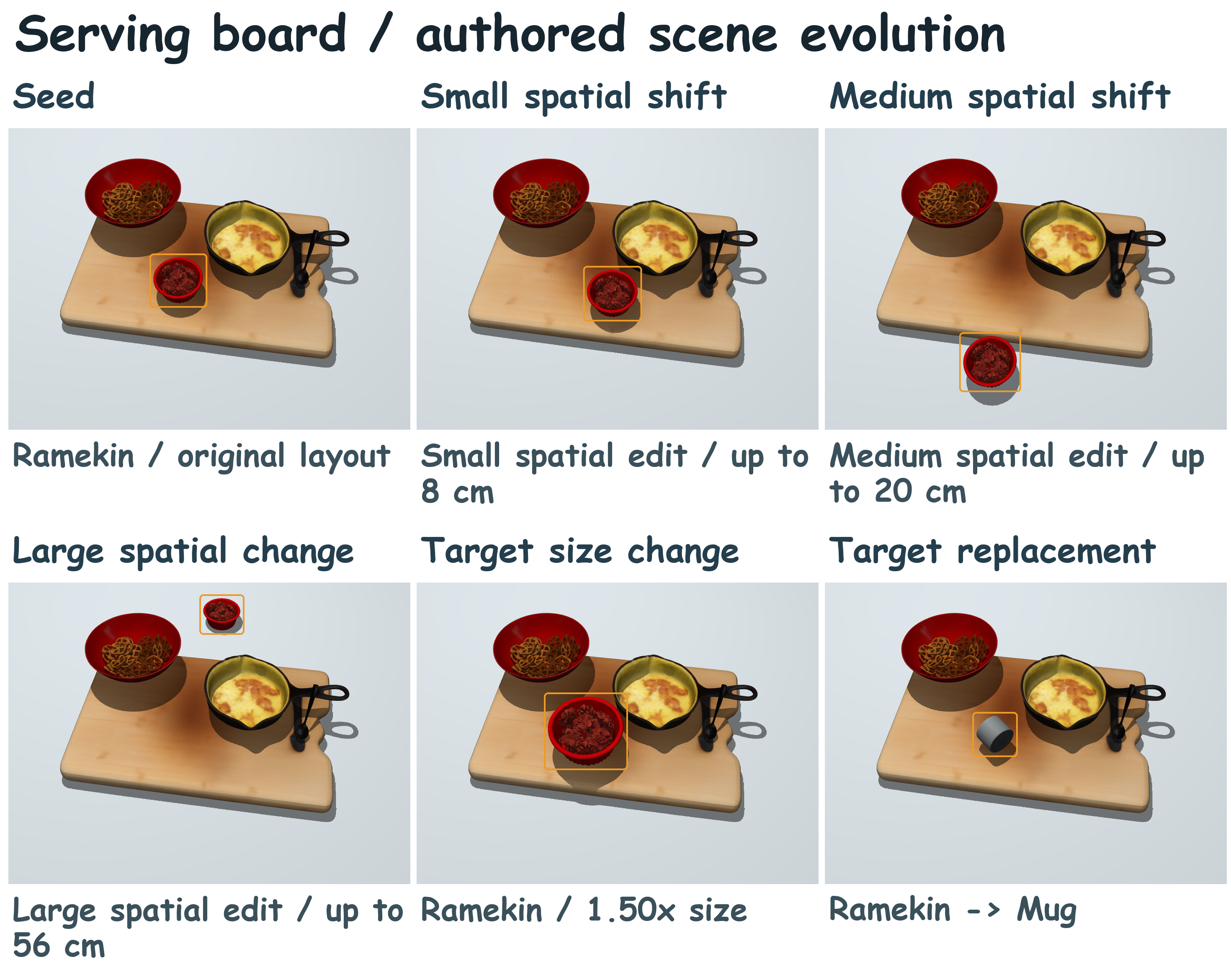}\\[4pt]
  \includegraphics[width=0.48\linewidth]{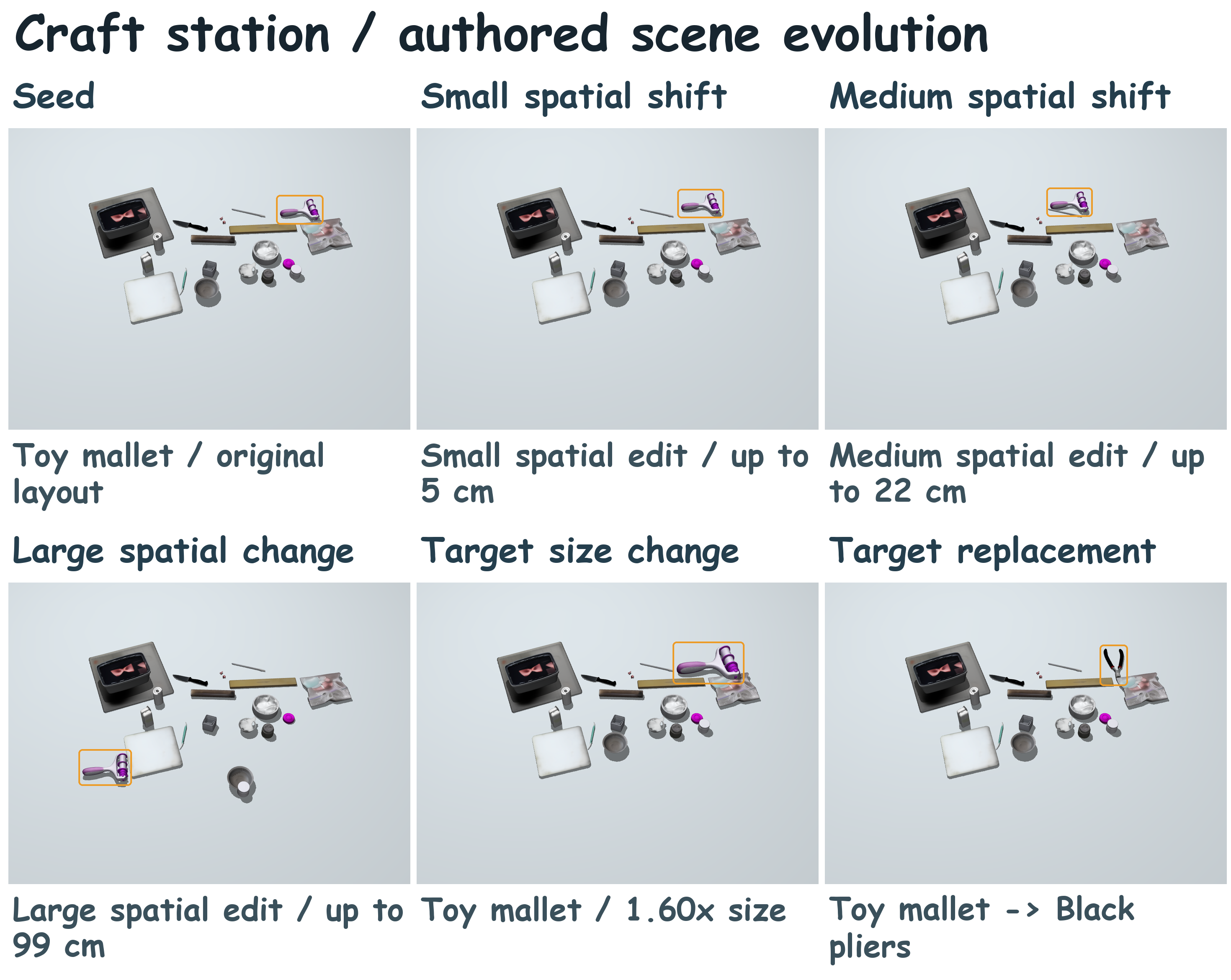}\hfill
  \includegraphics[width=0.48\linewidth]{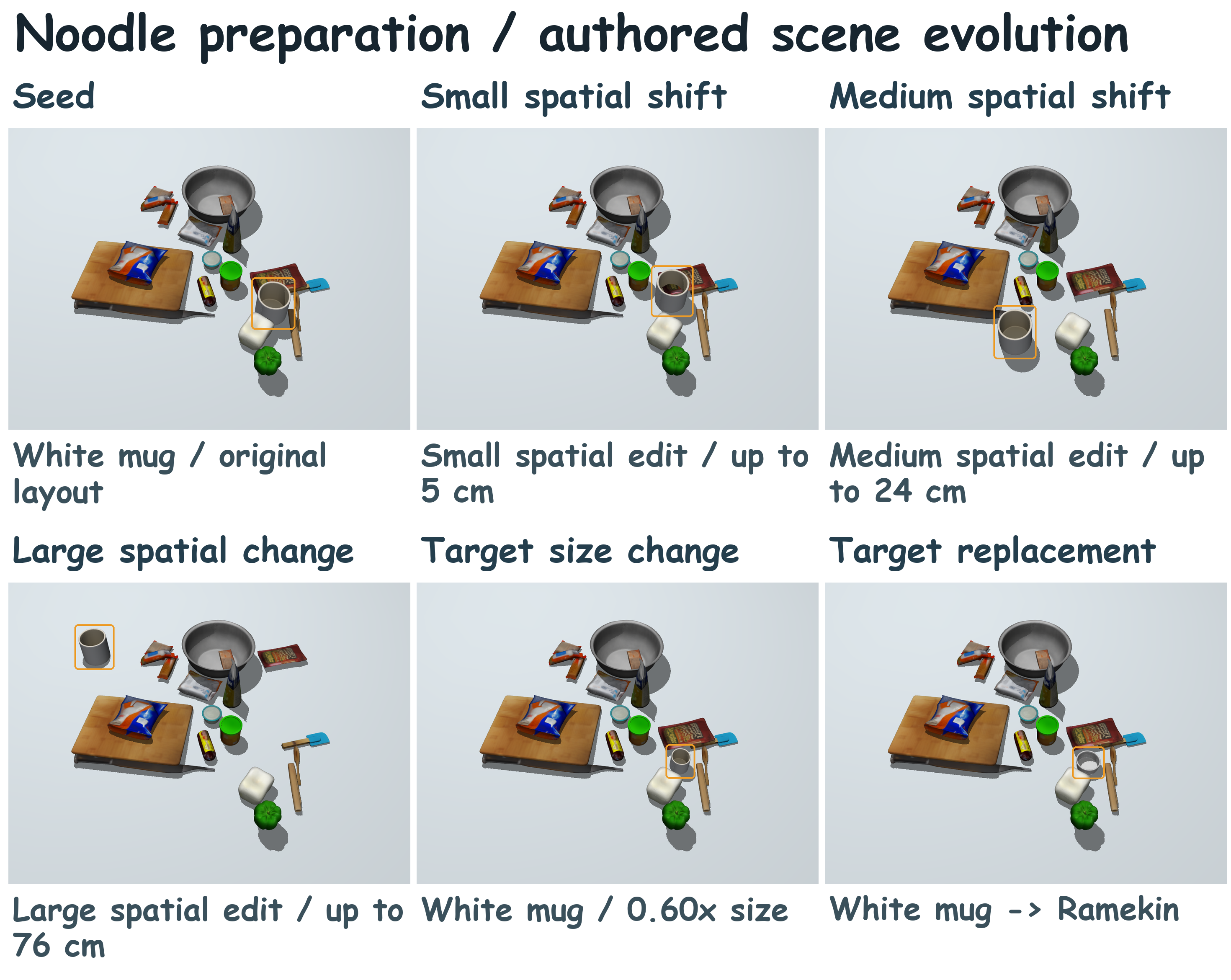}
  \vspace{-0.3cm}
  \caption{\textbf{Scene variations generated by \modelnamenc{}.}
Four examples show changes in object layout, size, and identity
for tool sorting, serving-board arrangement, craft-station
organization, and noodle preparation tasks.
Each panel contains a seed scene, three spatial perturbations
of increasing magnitude, a target-size change, and a
target-object replacement.
All displayed variants pass validity checks.\looseness-1}
  \label{fig:qual}
\end{figure*}

\begin{figure*}[t]
  \centering

  \includegraphics[height=9.6cm]{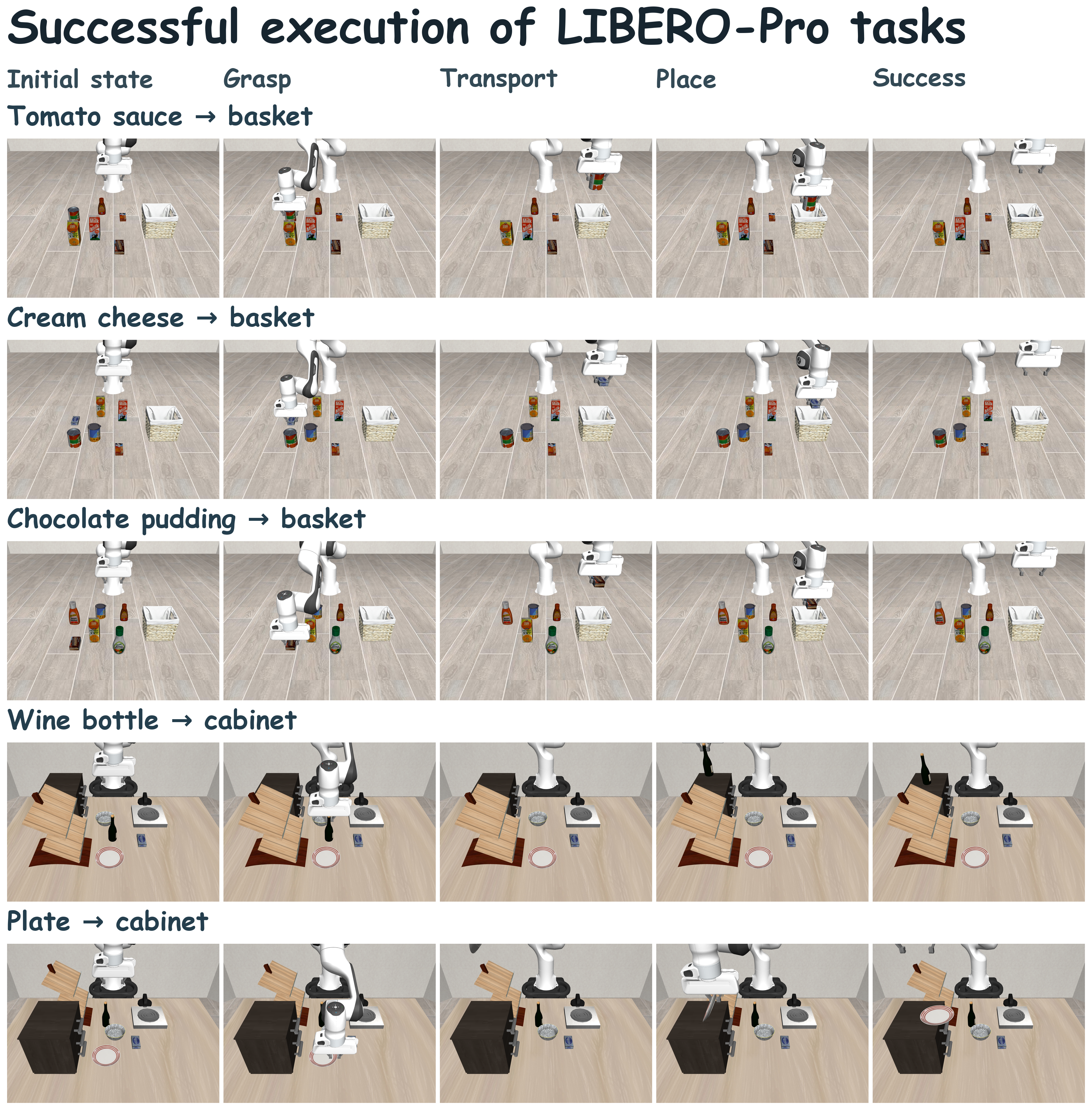}\hfill
  \includegraphics[height=9.6cm]{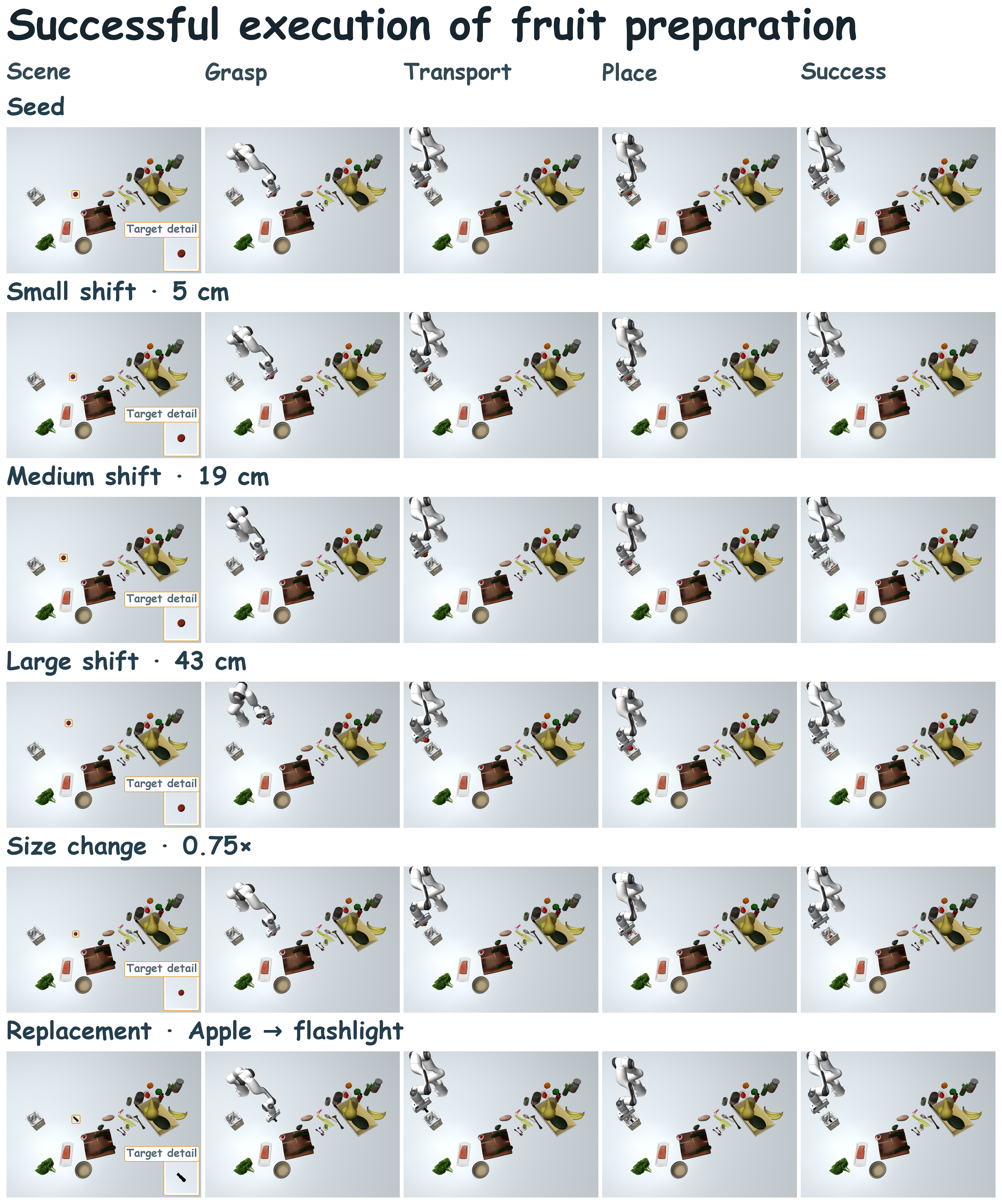}
   \vspace{-0.2cm}
  \caption{\textbf{Successful policy rollouts on benchmark
and generated environments.}
Left: five LIBERO-Pro tasks involving object placement
into a basket or cabinet.
Right: a fruit-preparation seed scene and five generated
variants, covering three spatial perturbations, target
resizing, and target replacement.
Each row shows one rollout, with columns depicting the
initial scene, grasp, transport, placement, and successful
task completion.\looseness=-1}
  \label{fig:qual:libero:more}
\end{figure*}

\textbf{Qualitative results.} 
\Cref{fig:qual} illustrates the scene variations produced
by the data engine from four seed scenes.
For each seed, we show three spatial perturbations of
increasing magnitude, a target-size change, and a
target-object replacement.
All displayed variants pass the scene-validity checks.
Additional successful policy rollouts on LIBERO-Pro tasks
and generated fruit preparation scenes are shown in
\Cref{fig:qual:libero:more}.

\flushcolsend
\section{Limitations}
\label{app:limitations}
\modelnamenc{} relies on foundation models for scene generation,
failure diagnosis, and potential compilation.
Errors in these models' understanding of objects, physical
interactions, or task requirements can affect the resulting
scenes and training signals.
Physical validity checks and control experiments
help detect such errors, but their coverage depends on the
checks and interventions used.
Results may also vary with changes to the underlying
models, making model versions and configurations important
for reproducibility.
The framework requires additional simulation and model
inference to generate environments, evaluate policy
rollouts, and validate diagnoses.
These costs constrain the number of rounds and task families
that can be explored under a fixed budget.
Our evaluation also focuses on simulated manipulation.
Transfer to physical robots remains to be established,
particularly for tasks sensitive to differences in contact
dynamics, sensing, and actuation.

\section{Broader Impact}
The ability to generate environments and training signals automatically has far-reaching
implications for robot learning, simulation, and embodied AI. \modelnamenc{} contributes to this
space by turning environment generation into a closed loop that responds to what a policy has
learned, which lowers the cost of producing diverse, verifiable training data. This can accelerate
the development of capable manipulation policies, improve the fidelity and coverage of simulated
benchmarks, and support scalable data generation for research groups without large teleoperation
budgets. While lowering the barrier to training capable robot policies raises potential concerns
about misuse and about biases inherited from the underlying foundation models, our framework is
designed for research use, judges success only through machine-checkable verifiers, and audits
every automated diagnosis with control experiments. We encourage responsible deployment practices
that keep such audits in place. Compared to collecting real-robot demonstrations at scale, the
approach is computationally efficient and requires no physical data collection, and we view its
benefits in controllable, verifiable data generation as outweighing its risks when applied
responsibly.

\end{document}